\documentclass{article}

\usepackage[final]{neurips_2025}

\usepackage[utf8]{inputenc} 
\usepackage[T1]{fontenc}    
\usepackage{hyperref}       
\usepackage{url}           
\usepackage{booktabs}       
\usepackage{amsfonts}       
\usepackage{nicefrac}       
\usepackage{microtype}      
\usepackage{xcolor}         
\usepackage{graphicx} 
\usepackage{tikz}
\usepackage{multirow}
\usepackage{tabularx}
\usepackage{array}
\usepackage{makecell}
\usepackage{xspace}
\usepackage{booktabs}
\usepackage{colortbl}
\newcommand{\best}[1]{\textbf{#1}}
\newcommand{\pmstd}[2]{#1 $\pm$ #2}
\newcommand{\up}{$\uparrow$}
\newcommand{\down}{$\downarrow$}
\definecolor{highlight}{RGB}{235,235,250}

\usepackage{makecell}
\usepackage{algorithm}
\usepackage{algorithmic}

\definecolor{thinkcolor}{rgb}{0.2, 0.4, 0.8}  
\definecolor{querycolor}{rgb}{0.1, 0.6, 0.2}  
\definecolor{summarycolor}{rgb}{0.9, 0.5, 0.1} 
\definecolor{answercolor}{rgb}{0.8, 0.2, 0.2}  
\definecolor{systemcolor}{rgb}{0.5, 0.5, 0.5} 
\definecolor{infocolor}{rgb}{0.6, 0.6, 0.6}  
\definecolor{searchbehaviorcolor}{HTML}{EFF9DA}
\definecolor{memorybehaviorcolor}{HTML}{F9EBDF}
\definecolor{multibehaviorcolor}{HTML}{CDF5F6}

\newcommand{\colortag}[2]{\textcolor{#1}{\texttt{#2}}}

\newcommand{\alg}{\textsc{MEM1}}

\hypersetup{
    colorlinks=true,
    linkcolor=blue,
    filecolor=magenta,      
    urlcolor=cyan,
    pdfpagemode=FullScreen,
    citecolor=violet,
}

\usepackage{amsmath}
\usepackage{amssymb}
\usepackage{mathtools}
\usepackage{amsthm}
\usepackage{enumitem}

\usepackage{caption}  
\usepackage{subcaption}
\usepackage{wrapfig}
\usepackage{soul}

\usepackage{tcolorbox}
\usepackage{colortbl}
\newtcolorbox[auto counter, number freestyle={\noexpand\arabic{\tcbcounter}}]{mycolorbox}[3][]{%
    fonttitle=\bfseries,
    title=#2~\thetcbcounter: #3,
    #1
}

\usepackage[capitalize]{cleveref}
\Crefname{figure}{Fig.}{Figs.}
\Crefname{table}{Tab.}{Tabs.}
\Crefname{algorithm}{Alg.}{Algs.}
\Crefname{section}{Sec.}{Secs.}
\Crefname{appendix}{App.}{Apps.}
\Crefname{proposition}{Prop.}{Props.}
\crefname{lemma}{Lemma}{Lemmas}
\crefname{observation}{Obs.}{Obs.}

\theoremstyle{definition}

\theoremstyle{definition}

\theoremstyle{remark}

\usepackage[textsize=tiny]{todonotes}

\let\citet\citep 

\def\eqref#1{Eq.~\ref{#1}}

\def\1{\bm{1}}

\DeclareMathAlphabet{\mathsfit}{\encodingdefault}{\sfdefault}{m}{sl}
\SetMathAlphabet{\mathsfit}{bold}{\encodingdefault}{\sfdefault}{bx}{n}

\newcommand{\ie}{\textit{i.e.,}\xspace}
\newcommand{\eg}{\textit{e.g.,}\xspace}

\DeclarePairedDelimiterX\braket[2]{\langle}{\rangle}{#1\,\delimsize\vert\,\mathopen{}#2}

\newcommand{\squishlisttwo}{
 \begin{list}{$\bullet$}
  { \setlength{\itemsep}{1pt}
     \setlength{\parsep}{0pt}
    \setlength{\topsep}{0pt}
    \setlength{\partopsep}{0pt}
    \setlength{\leftmargin}{1em}
    \setlength{\labelwidth}{1.5em}
    \setlength{\labelsep}{0.5em} } }

\newcommand{\squishend}{
  \end{list}  }

\newcommand{\el}{\end{flushleft}}
\newcommand{\bl}{\begin{flushleft}}

\theoremstyle{definition}

\setcitestyle{numbers,square,comma}

\title{\alg{}: Learning to Synergize Memory and Reasoning for Efficient Long-Horizon Agents}

\author{%
  Zijian Zhou\textsuperscript{*12}
  \quad
  Ao Qu\textsuperscript{*13}
  \quad
  Zhaoxuan Wu$^{1}$
  \quad
  Sunghwan Kim$^{4}$ \\
  \textbf{Alok Prakash}$^1$
  \quad
  \textbf{Daniela Rus}$^{13}$
  \quad
  \textbf{Jinhua Zhao}$^{13}$
  \quad
  \textbf{Bryan Kian Hsiang Low}$^{13}$ \\
  \textbf{Paul Pu Liang}$^{3}$ \\
  $^{1}$Singapore-MIT Alliance for Research and Technology Centre \\
  $^{2}$National University of Singapore \\
  $^{3}$MIT \\
  $^{4}$Yonsei University \\
}

\begin{document}
\makeatletter
\def\blankfootnote{\xdef\@thefnmark{}\@footnotetext}
\makeatother

\maketitle

\begin{abstract}
Modern language agents must operate over long-horizon, multi-turn interactions, where they retrieve external information, adapt to observations, and answer interdependent queries. Yet, most LLM systems rely on full-context prompting, appending all past turns regardless of their relevance. This leads to unbounded memory growth, increased computational costs, and degraded reasoning performance on out-of-distribution input lengths. We introduce \textbf{\alg{}}, an end-to-end reinforcement learning framework that enables agents to operate with \textbf{constant memory} across long multi-turn tasks. At each turn, \alg{} updates a \textbf{compact shared internal state} that jointly supports memory consolidation and reasoning. This state integrates prior memory with new observations from the environment while strategically discarding irrelevant or redundant information. To support training in more realistic and compositional settings, we propose a simple yet effective and scalable approach to constructing multi-turn environments by composing existing datasets into arbitrarily complex task sequences. Experiments across three domains, including internal retrieval QA, open-domain web QA, and multi-turn web shopping, show that \alg{}-7B improves performance by $3.5\times$ while reducing memory usage by $3.7\times$ compared to Qwen2.5-14B-Instruct on a \textbf{$16$-objective} multi-hop QA task, and \textbf{generalizes beyond the training horizon}. Our results demonstrate the promise of reasoning-driven memory consolidation as a scalable alternative to existing solutions for training long-horizon interactive agents, where both efficiency and performance are optimized. Code can be found at \url{https://github.com/MIT-MI/MEM1}.

\blankfootnote{\textbf{*} Equal contribution. Correspondence: \texttt{zhou\_zijian@u.nus.edu, qua@mit.edu}}
\end{abstract}
\section{Introduction}
\label{sec:intro}

Large language models (LLMs) have shown remarkable performance in single-turn tasks such as question answering, summarization, and code generation~\citep{brown2020language,touvron2023llama,anthropic2024claude3}. However, emerging real-world applications increasingly operate over multiple turns\textemdash searching documents, interacting with environments~\citep{zhouwebarena}, and making decisions based on evolving external information~\citep{wang2024survey}. Examples include research agents such as OpenAI and Gemini Deep Research~\citep{openai2025deepresearch,google2024deepresearch} that automate complex tasks by iteratively gathering information, and web-navigation agents such as OpenManus~\citep{openmanus2025} and BrowserUse~\citep{browser_use2024}, which must complete goals across dozens of interactive turns. 

Unlike traditional tasks where the input is static or self-contained, long-horizon settings often involve answering a sequence of related questions, requiring the agent to continuously retrieve new information, revise beliefs, and adapt to evolving contexts over time.
For instance, consider a research assistant tasked with ``What’s the evidence for X?''.
Subsequent queries like ``Who published it?'' require further information retrieval, while “Is the source credible?” calls for self-reflection and assessment.
Each query builds on the previously collected and accumulated information. 
Similarly, a shopping assistant may be first asked ``Which product is cheapest?'', then ``What are its reviews?'', and ``Is it compatible with my device?''.
These interactions span multiple turns, featuring evolving contexts and compound reasoning. 


In systems designed for long-horizon settings, a common approach is to append all past observations, actions, and thoughts to the context at every turn~\citep{wei2022chain,yao2023react}.
This forces the model to operate with an unboundedly growing context, which introduces three key challenges. 
\textbf{(1) Growing inference cost and memory usage}.  Transformer-based LLMs typically incur $O(N^2)$ compute cost (or $O(N)$ with Key-Value caching) and $O(N)$ memory usage as the context length $N$ increases~\citep{vaswani2017attention}. 
Consequently, deploying these models requires reserving large GPU memory on modern inference frameworks to accommodate the growing context~\citep{kwon2023efficient,sglang}, often leading to significant wastage of computing resources.
\textbf{(2) Generalization limits beyond the training horizon}. Ongoing conversations with context length exceeding that in the training data become out-of-distribution for the model. The model struggles to manage and reason over such unfamiliar long-horizon inputs~\citep{yoon-etal-2024-compact}. 
\textbf{(3) Overloaded and inefficient context}. The accumulation of irrelevant or redundant content dilutes the model’s attention. This reduces its ability to reason effectively, even when relevant information is technically still present within the prompt~\citep{an2025effective,liu-etal-2024-lost,wu2025longmemeval}.


Recent progress in long-context modeling largely targets static inputs (\eg long documents) and does not address multi-turn interaction with external environments~\citep{beltagy2020longformer,gu2025attention}. Some other approaches introduce external memory modules (\eg summarizers or retrievers)~\citep{yoon-etal-2024-compact,li2023compressing,chhikara2025mem0,xu2025mem}, but these are typically trained separately and cannot be optimized end-to-end with the agent’s policy. This also introduces additional engineering overhead, as engineers must manage and integrate two separate models. Meanwhile, existing works on tool-using agent systems trained with reinforcement learning leave memory management unsolved, letting the prompt length grow unboundedly~\citep{jin2025search,zheng2025deepresearcher}. A natural question is raised: \textit{ \textbf{Can a language model learn to consolidate its memory as part of its reasoning process} so that it retains only what is essential for solving the task?}


\begin{figure}[t]
    \centering
    \includegraphics[width=\linewidth]{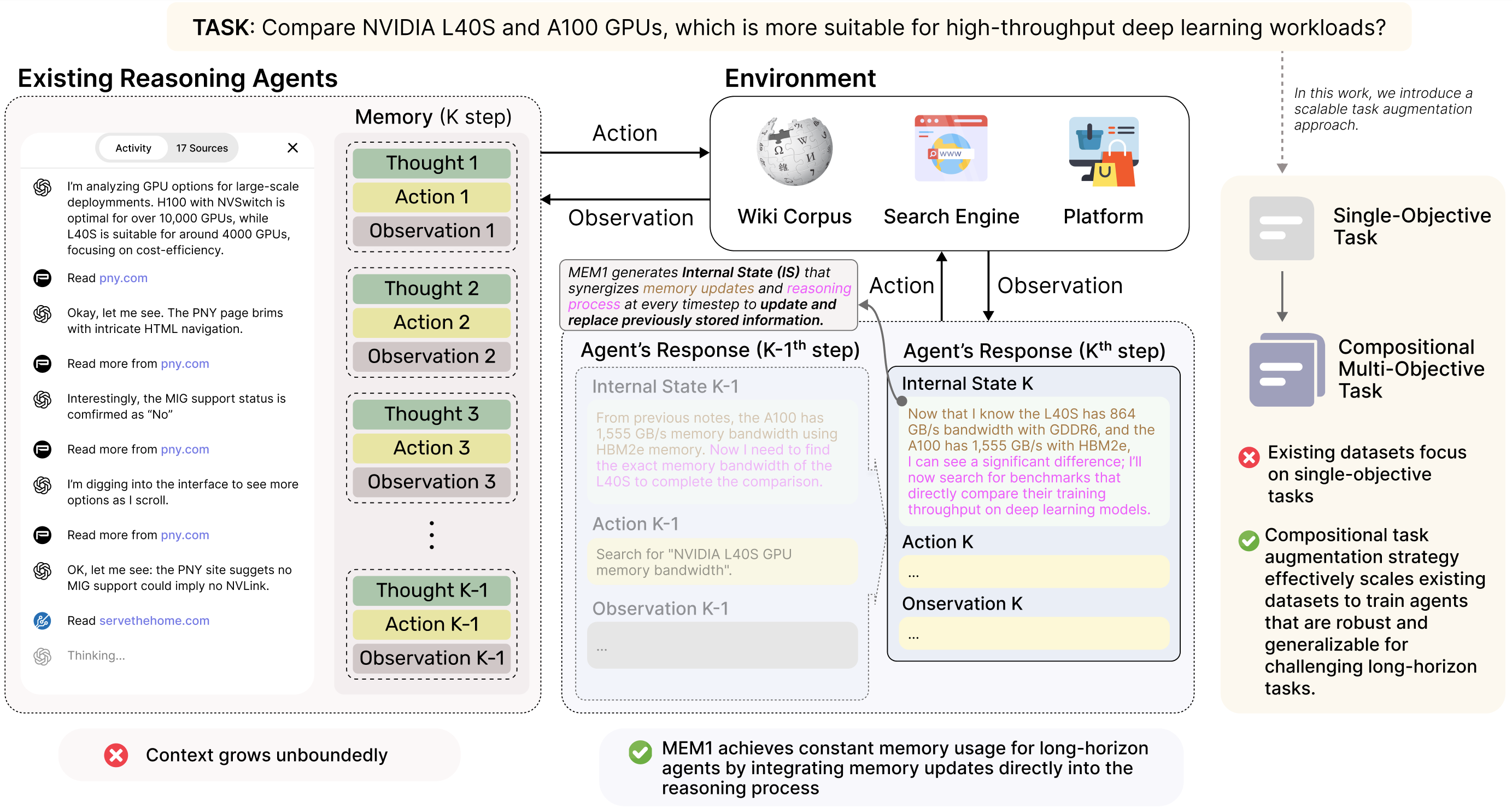}
    \caption{Comparison of memory management between \alg{} and existing reasoning agents. While existing agents for long-horizon tasks \cite{jin2025search,yuan2025agent,zheng2025deepresearcher} continuously append thoughts (typically enclosed in <think></think>), actions, and observations, resulting in an ever-growing context, our \alg{} agent learns to keep updating an internal state (enclosed in <IS></IS>) that blends thought and memory, discarding the contents from previous steps to achieve constant memory usage during the task. On the other hand, while existing environments and datasets focus on single-objective tasks, our task augmentation method effectively scales up these tasks to enable long-horizon agent training.}
    \label{fig:flow}
    \vspace{-5mm}
\end{figure}

Motivated by this question, we present \alg{}: \textbf{M}emory-\textbf{E}fficient \textbf{M}echanism via learning \textbf{1}-step integrated reasoning and consolidation\textemdash a method for training LLM agents that maintain constant memory usage across arbitrarily long horizons.
\begin{wrapfigure}{r}{0.4\textwidth}  
    \centering
    \includegraphics[width=0.38\textwidth]{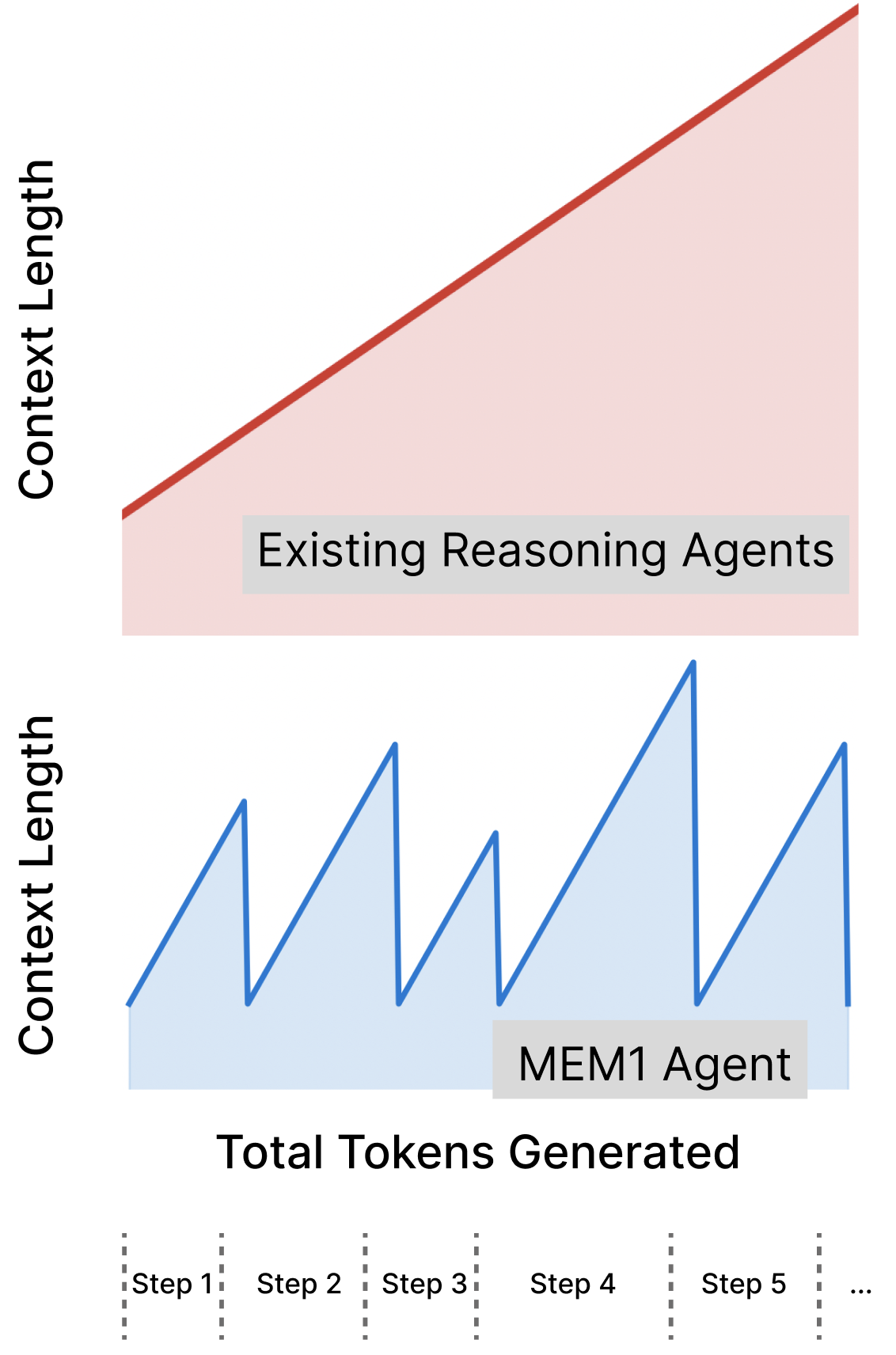}
    \caption{A conceptual comparison of context length between the \textcolor{blue}{\alg{} agent} and \textcolor{red}{existing reasoning agents} when handling long-horizon tasks. Our agent learns to discard the previous context (except for the prompt and initial query) immediately after generating a new internal state and action, resulting in near-constant memory usage.}
    \label{fig:token_comparison}
    \vspace{-2mm}
\end{wrapfigure}
As illustrated in~\cref{fig:flow}, at each turn, the model updates a consolidated state composed of prior memory and newly obtained information. This consolidated state becomes the agent’s only retained memory, allowing all external tool outputs to be discarded after use, which prevents prompt expansion altogether as illustrated by~\cref{fig:token_comparison}. A key insight of our method is that inference-time reasoning~\citep{wei2022chain,deepseek2025r1,muennighoff2025s1,yue2025inference} serves a dual function: it not only provides deeper insight into the current query but also acts as a form of ``working memory''~\citep{baddeley1974working}, extracting key components from gathered information to build an evolving understanding. By unifying reasoning and memory consolidation, \alg{} offers an elegant solution where the agent learns to reason and remember within a shared representational space without requiring additional modules or architectural changes.

We train this behavior end-to-end with reinforcement learning (RL)~\citep{sutton2018reinforcement,pmlr-v202-zhu23f}, optimizing for task success via verifiable rewards~\citep{deepseek-math}. Although not explicitly optimized for memory efficiency through reward signals, the agent learns to manage memory as part of its policy, resulting in near-constant memory usage across long horizons. Additionally, we notice that current training and evaluation environments predominantly focus on single-objective tasks~\cite{kwiatkowski-etal-2019-natural,yang-etal-2018-hotpotqa,press-etal-2023-measuring}, limiting their ability to fully prepare agents for realistic, long-horizon scenarios that inherently involve multiple sequential objectives. To address this challenge, we introduce a scalable task augmentation approach, transforming existing single-objective QA datasets into complex multi-objective tasks through compositions of $N$ multi-hop questions. 
This approach enables us to repurpose standard benchmarks in our community to more effectively train and evaluate agents on long-horizon reasoning, an increasingly important capability in real-world applications.

To evaluate our method comprehensively, we employ diverse multi-turn environments, including internal retrieval-augmented QA~\citep{kwiatkowski-etal-2019-natural,yang-etal-2018-hotpotqa}, open-domain Web QA~\citep{zheng2025deepresearcher}, and complex multi-turn agent shopping scenarios in WebShop~\citep{yao2022webshop}. Across these scenarios, \alg{} consistently rivals the performance of leading baselines while delivering efficiency gains up to $3.5\times$ in memory usage. Moreover, agents trained on our augmented 2-objective compositions generalize effectively to tasks involving up to 16-objective compositions. 
Notably, at the 16-objective level, our \alg{} achieves superior accuracy compared to all baseline methods, along with $1.27\times$ lower peak memory usage and $1.78\times$ faster inference compared to the respective best uncollapsed baseline.

\section{Related Work}
\paragraph{LLM agents in multi-turn environment.}
LLM-based agents have evolved from handling single-turn queries to serving as autonomous agents capable of multi-turn interactions such as web navigation~\citep{yao2022webshop, zhouwebarena} and complex research~\citep{zheng2025deepresearcher}.
To enable such capabilities, Yao et al.~\citep{yao2023react} introduced the ReAct (\ie Reason + Act) framework, which enhances LLMs' ability to interact with external environments by interleaving reasoning and action. Building on this reasoning-acting prompting paradigm, subsequent works have explored ways to improve agent performance through natural language feedback, enabling iterative refinement~\citep{shinn2023reflexion, madaan2023self}. Recently, inference-time scaling has emerged as a promising direction for enabling complex reasoning, with prior research incorporating evaluators (\eg verifier, reward model)~\citep{snell2024scaling, liu2025inference} or world models~\citep{chae2024web}.
In addition, there are two major lines of training approaches: (1) behavior cloning (BC), which involves imitating expert trajectories to guide agent behavior by supervised fine-tuning (SFT)~\citep{yin2023lumos, deng2023mind2web, cheng2024seeclick}, and (2) reinforcement learning (RL), which optimizes agent policies by incentivizing desirable outcomes through rewards~\citep{song2024trial, bai2024digirl, qi2024webrl}. These methods aim to align the agents' behaviors with task objectives, enabling more robust and generalizable performance.


\paragraph{Memory management for LLM agents.}
A widely adopted approach to memory management in LLM-based agent systems involves appending all prior information, such as observations, intermediate thoughts, and actions, into the prompt at each interaction turn~\citep{yao2023react}. While this method is straightforward and effective when the number of interactions required is small, it results in unbounded context growth, leading to linearly scaled inference memory. Moreover, long contexts often contain irrelevant or redundant information, which impairs the model’s reasoning capabilities~\citep{an2025effective, liu-etal-2024-lost, wu2025longmemeval}. To mitigate these issues, recent studies have proposed external memory frameworks, including retrieval-augmented generation and summarization modules~\citep{yoon-etal-2024-compact, li2023compressing, chhikara2025mem0, xu2025mem}. However, these methods are typically trained or used independently of the agent’s policy, creating a disconnect between memory and the reasoning process. In addition, managing and integrating such modules often incurs extra computational overhead and system complexity. Despite these advancements, many RL approaches for training LLM agents still rely on accumulating the full interaction history as memory~\citep{jin2025search, zheng2025deepresearcher, qi2024webrl}, leaving memory management during training an underexplored area. In this work, we seek to bridge this gap by tightly integrating memory with the agent’s reasoning process, thereby enabling more efficient and context-aware decision-making.
\section{\alg{}}
\label{sec:methodology}

Complex reasoning tasks often require an iterative process of information gathering and synthesis, as seen in applications such as ``deep research''~\citep{openai2025deepresearch, jiang-etal-2024-unknown} and web-based agents~\citep{nakano2022webgpt, gur2024webagent}. Recent advances in agent design involve interaction loops that interleave chain-of-thought reasoning~\citep{wei2022chain, deepseek2025r1}, environment interaction, and real-world feedback collection. To explicitly capture these core elements, we annotate each component using XML-style tags: $\colortag{thinkcolor}{\texttt{<IS>}}$ for internal state (reasoning), $\colortag{querycolor}{\texttt{<query>}}$ for environment queries, $\colortag{answercolor}{\texttt{<answer>}}$ for the agent’s responses, and $\colortag{infocolor}{\texttt{<info>}}$ for external observations or tool outputs. \alg{} adopts a \textbf{learned} approach to iterative state updating and consolidation, ensuring that only the most recent set of $\colortag{thinkcolor}{\texttt{<IS>}}$, $\colortag{querycolor}{\texttt{<query>}}$, $\colortag{answercolor}{\texttt{<answer>}}$, and $\colortag{infocolor}{\texttt{<info>}}$ elements is retained in the prompt at any given time. This design maintains a bounded and semantically relevant context, promoting efficient and coherent multi-step reasoning.

\subsection{Memory as Part of Reasoning}

\begin{figure}
    \centering
    \includegraphics[width=0.99\linewidth]{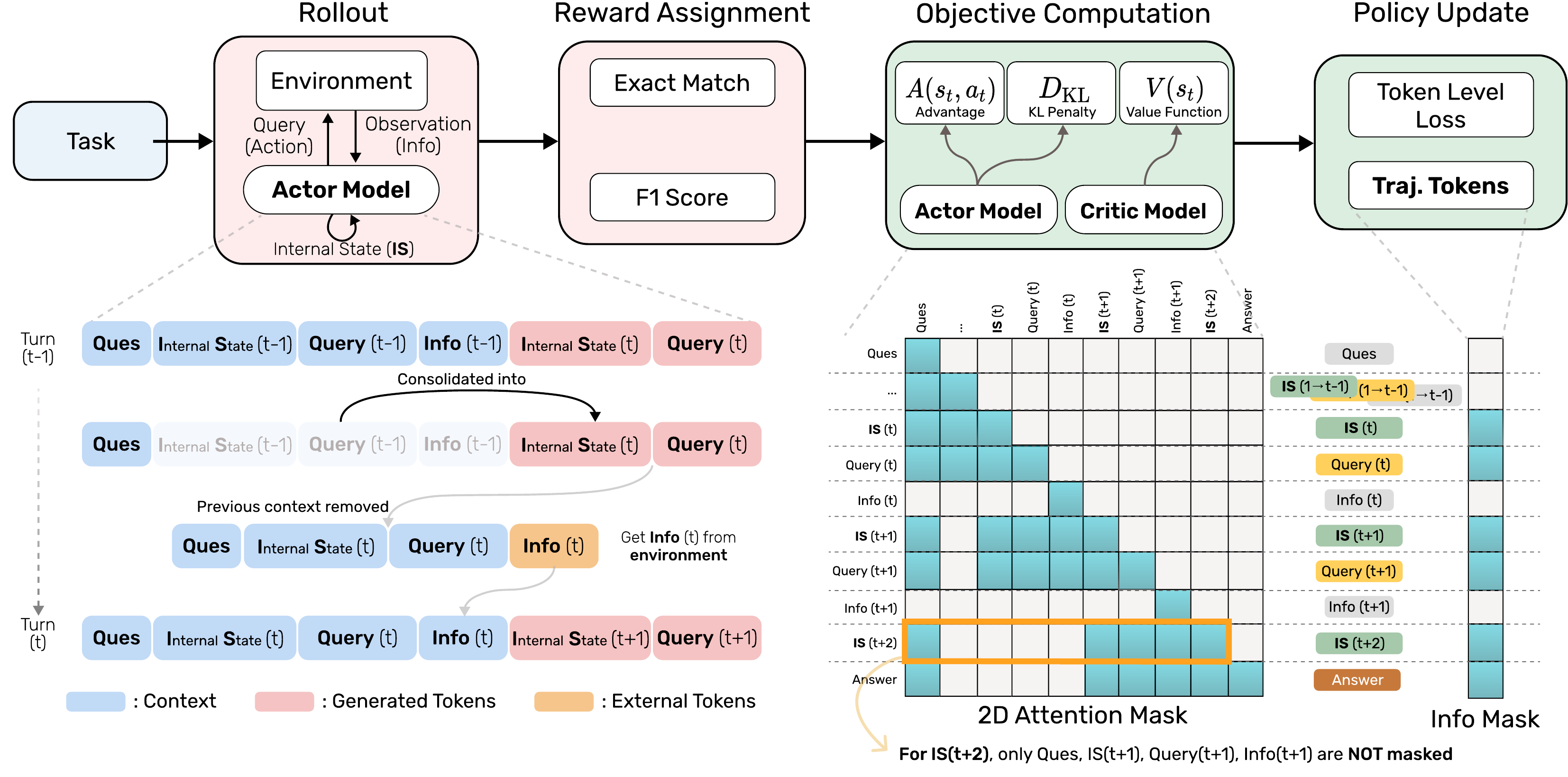}
    \caption{(Top): the RL pipeline used to train \alg{}. (Bottom left): The evolution of context in \alg{}--old $\colortag{thinkcolor}{<IS>}$, $\colortag{querycolor}{<query>}$, $\colortag{infocolor}{<info>}$ are cleared as new states enter the context. The mechanism is used in the rollout. (Bottom right): the $2$D attention mask used during the objective computation stage. The mask is applied during the forward pass to compute action log-probabilities for the actor model and state value estimates for the critic model. During the policy update stage, the information mask is then applied to the full trajectory, masking out tokens that were not generated by the model itself.}
    \label{fig:mem1-mechanism}
\end{figure}

To achieve a constant memory, \alg{} is particularly trained to iteratively refine its understanding by processing new information in conjunction with a consolidation of its prior state. 
At each turn $t$, the agent produces a new $\colortag{thinkcolor}{\texttt{<IS\_t>}}$ element, which summarizes past information and reasons about subsequent actions. 
Following this, the agent generates an action—either a $\colortag{querycolor}{\texttt{<query\_t>}}$ to interact with the environment, or an $\colortag{answercolor}{\texttt{<answer\_t>}}$ if a direct response is warranted. 
If the agent issues a $\colortag{querycolor}{\texttt{<query\_t>}}$, the corresponding feedback from the environment is appended as $\colortag{infocolor}{\texttt{<info\_t>}}$. 
At the next turn, $t+1$, the agent consolidates the tuple $\big(\colortag{thinkcolor}{\texttt{<IS\_t>}}, \colortag{querycolor}{\texttt{<query\_t>}}, \colortag{infocolor}{\texttt{<info\_t>}}\big)$ into a new $\colortag{thinkcolor}{\texttt{<IS\_(t+1)>}}$, which serves as the basis for further interactions. 
After each turn, all tags from the previous turn $t$ are pruned from the context, effectively compressing memory and preventing prompt bloat. 
\cref{fig:mem1-mechanism} (bottom left) illustrates the evolution of the model’s context over time. 
At any given turn, the agent retains at most two $\colortag{thinkcolor}{\texttt{<IS>}}$ elements, two $\colortag{querycolor}{\texttt{<query>}}$ elements, and one $\colortag{infocolor}{\texttt{<info>}}$ element, ensuring bounded and efficient memory usage.
The detailed rollout algorithm is in~\cref{alg:mem1-rollout} of~\cref{app:algo}.

RL offers a powerful mechanism for shaping agent behavior through reward signals~\citep{sutton2000policy}. In \alg{}, we leverage this framework to incentivize effective state consolidation by designing environments in which the agent is rewarded only when it strategically retains and integrates useful information. Specifically, we construct tasks that require numerous interactions with the environment to arrive at a correct answer (see \cref{sec:task-design}). Success depends on the agent’s ability to rely on information collected along the inference path. At each turn, we prune the agent’s context to retain only the most recent $\colortag{thinkcolor}{\texttt{<IS>}}$, forcing the agent to perform memory consolidation as part of its reasoning process. Without access to full historical context, the agent must learn to preserve and update relevant knowledge internally in order to reap the reward. This learning procedure mirrors how humans cultivate memorization skills through structured tasks such as Sudoku or crosswords~\citep{cognitiveconnection2024shortterm}, where success hinges on selectively attending to key information and building upon it. Over time, such tasks help individuals develop cognitive strategies that jointly support efficient memorization and reasoning, similar to our RL method for training \alg{}.

\subsection{Masked Trajectory for Policy Optimization}

In the previous section, we detailed the rollout process of our RL pipeline. 
However, unlike conventional multi-turn agents that preserve a static context during generation, \alg{} introduces a unique challenge: its mechanism continuously updates the context at each turn by consolidating prior memory and pruning irrelevant tokens. 
This dynamic context update disrupts the continuity of the token generation trajectory, complicating the estimation of token-wise advantages in policy optimization algorithms such as PPO and Reinforce++~\citep{schulman2017proximal,hu2025reinforcepp}, where trajectories are typically assumed to be linear.

To address this, we introduce a masked trajectory approach that reconstructs a logically coherent full trajectory by stitching together multiple interaction turns with evolving contexts.
This unified trajectory mimics a standard multi-turn rollout and comprises a sequence of tuples $\tau_t = (\colortag{thinkcolor}{\texttt{<IS\_t>}}, \colortag{querycolor}{\texttt{<query\_t>}}, \colortag{infocolor}{\texttt{<info\_t>}})$ for $t \in [1, T-1]$, where $T$ denotes the total number of interaction turns. 
The $T$-th turn outputs the final answer $\tau_T = (\colortag{thinkcolor}{\texttt{<IS\_t>}}, \colortag{answercolor}{\texttt{<answer\_t>}})$.
The full trajectory encodes all information needed for accurate policy learning while respecting \alg{}'s memory consolidation at each turn. 
\cref{fig:mem1-mechanism} (bottom left) demonstrates the evolution of the agent's context.

To ensure that policy gradients are correctly computed under this consolidated memory regime, we apply a \textbf{two-dimensional attention mask}~\citep{ruslan2024_4d_masks} across the full trajectory. 
This mask restricts each token’s attention to only the tokens retained in memory at the time that token was generated. 
Specifically, for a token position $k$, we mask out all prior tokens that are not part of the consolidated memory corresponding to the context at that turn. 
This masking strategy enables accurate computation of the policy objective: letting $s_k$ denote the masked input state and $a_k$ the predicted token, the log-probability ratio $\rho_k(\theta) = \frac{\pi_\theta(a_k \mid s_k)}{\pi_{\theta_{\text{old}}}(a_k \mid s_k)}$ remains valid and tractable, in turn ensuring that the advantage, KL penalty, and value estimation are correct. A visualization is in \cref{fig:mem1-mechanism} (bottom right). Furthermore, following~\citet{jin2025search}, we incorporate a one-dimensional attention mask over retrieved external information during policy updates for both the actor and critic networks. 
This ensures that gradient updates are localized to only tokens generated by the agent. 
\Cref{fig:mem1-mechanism} (bottom right) shows the masking mechanism that enables stable and accurate policy optimization under \alg{}’s memory-constrained execution.

\subsection{Multi-Objective Task Design}
\label{sec:task-design}

Although our proposed method is designed to address the critical challenges of agentic multi-turn interaction with the external world, there are limited publicly available datasets that support training for such long-horizon interactive processes. Existing benchmarks, such as HotpotQA~\citep{yang-etal-2018-hotpotqa}, Bamboogle~\citep{press-etal-2023-measuring}, and 2wiki~\citep{ho-etal-2020-constructing}, are often cited as multi-hop benchmarks, yet they typically involve only two information-seeking steps. Moreover, these datasets are not explicitly structured to support long-horizon interactions that necessitate the agent to manage the memory state.

To bridge this gap, we introduce a novel task---multi-objective question \& answering (QA)---that extends the number of reasoning steps required to solve a problem. Building on existing multi-turn datasets such as HotpotQA and Natural Question~\citep{yang-etal-2018-hotpotqa,kwiatkowski-etal-2019-natural}, we interleave multiple questions from the original QA corpus and construct a single composite query that requires answering all constituent sub-questions, shown in Prompt~\ref{multi-prompt} of~\cref{app:prompts}. This formulation compels the agent to perform multiple search queries, each targeting a distinct sub-question or sub-objective, and then integrate the retrieved answers to form a comprehensive final response. Compared to the original tasks, our synthesized multi-objective, multi-hop setting significantly increases the number of search and reasoning turns required, leading to more complex, memory-intensive interactions.

\label{subsec:implementation}
\section{Experiments \& Results}

We empirically demonstrate the effectiveness of our approach in training the \alg{} agent to perform multi-turn tasks while preserving a near-constant-sized memory state. We evaluate \alg{} against several baselines using a comprehensive set of metrics categorized into \textit{accuracy} (\eg Exact Match, F1 score, Environment Reward) and \textit{efficiency} (e.g., Peak Token Usage, Dependency Length, Inference Time). All \alg{} variants are fine-tuned from the Qwen2.5-7B Base model~\citep{qwen2.5}. We use PPO~\citep{schulman2017proximal} as the RL algorithm as it computes token-level advantages, bringing stability to the training process. While we also experimented with instruction-tuned and supervised fine-tuned models using curated high-quality trajectories, reinforcement learning from the base model consistently yielded the best performance and generalization.

Our experiments are conducted in two standard environments, each reflecting real-world scenarios that require multi-turn agent interactions. The first environment is question answering with retrieval-augmented generation (RAG)~\citep{kwiatkowski-etal-2019-natural,yang-etal-2018-hotpotqa}, where the agent must answer queries by retrieving relevant information from an external knowledge store (either a database or an online search engine). We trained on RAG with a local database (\ie Wikipedia Corpus) and evaluated on tasks involving open web browsing. 
For QA, following~\cref{sec:task-design}, we construct multi-objective tasks and tested the model performance on tasks with more questions than seen in the training.  
The second environment is WebShop navigation~\citep{yao2022webshop}, where the agent assists users in online shopping by browsing a website and selecting items based on natural language descriptions. This task requires the agent to iteratively read page content and make navigation decisions, following protocols similar to those in WebGPT~\citep{nakano2022webgpt}.





\subsection{Implementation Details}
\label{sec:exp-impl-details}

\paragraph{Datasets and evaluation metrics.} We train two versions of \alg{} agent for both long-horizon QA and web navigation. For long-horizon QA, we augment the multi-hop QA dataset from~\citep{jin2025search} that mixes data from both HotpotQA~\citep{yang-etal-2018-hotpotqa} and Natural Question~\citep{kwiatkowski-etal-2019-natural} to form a $2$-objective composite task. 

For the web agent, we use the WebShop environment~\citep{yao2022webshop}, which also produces a reward during training~\citep{yuan2025agent}. 
For all datasets, the train-test split follows the original papers. 
During RL training, we employ the exact match (EM) metric for QA tasks (details in~\cref{app:metrics}) and the environment reward for WebShop~\citep{yao2022webshop,yuan2025agent}.
To evaluate the effectiveness of various approaches, we measure the EM and F1 score for QA tasks and final reward for the WebShop environment~\citep{yao2022webshop,yuan2025agent}. 
To evaluate the efficiency, we consider the peak token usage, average dependency, and average inference time. The test datasets are obtained from the original papers which consist of out-of-distribution data.
The former two metrics measure the memory efficiency, while the latter measures the time efficiency. 
The detailed definitions of the metrics are in~\cref{app:metrics}.
The prompt and format can be found in~\cref{app:prompts}. 

\paragraph{Baselines.} To evaluate the accuracy and efficiency of \alg{}, we compare it against a diverse set of baselines designed to either enhance task performance or manage context effectively. For the QA environment, we benchmark accuracy against Search-R1 \citep{jin2025search}, DeepResearcher \citep{zheng2025deepresearcher}, and a larger-scale model, Qwen2.5-14B-Instruct \citep{qwen2.5}. Details about Search-R1 and DeepResearcher can be found in~\cref{app:baselines}. For the WebShop environment, we compare against Agent-FLAN \citep{chen-etal-2024-agent}, Agent-R \citep{yuan2025agent}, and AgentLM \citep{zeng2023agenttuning}. To assess efficiency, we consider two context compression baselines using models of the same parameter size as \alg{}. First, we apply \alg{}’s agentic \textit{truncation} prompt template and rollout to a standard instruct model, isolating the benefits of prompt and rollout design alone. Second, we evaluate A-MEM~\citep{xu2025mem}, which augments an Instruct model with a vector database for memory retrieval, capturing the effect of external memory modules in agentic systems. We additionally train a supervised fine-tuned (SFT) model using trajectories curated from GPT-4o~\citep{openai2024gpt4o} based on \alg{}'s rollout and compare it with the RL-trained agent.



\paragraph{Meta info injection.} In our agentic pipeline, the agent's context is programmatically truncated at each turn—immediately after it generates a search query or an answer—following the procedure outlined in \cref{sec:methodology}. As past context is truncated, the agent may have difficulty determining when to terminate. To address this, we prepend a hint [\texttt{HINT: YOU HAVE \{turns\_left\} TURNS LEFT}] at the beginning of each $\colortag{infocolor}{\texttt{<info>}}$ tag to remind the agent of its remaining turns budget. For all experiments, we set the maximally allowed turns to $6$ for $1$-objective to $4$-objective tasks and $20$ for more difficult tasks to avoid excessively long trajectories.





\subsection{\alg{} on Multi-Objective Multi-Hop Tasks}\label{sec:exp-impl-details:multi-obj}

One key advantage of \alg{} agents lies in their efficient management of long-horizon interactions with the environment.
To demonstrate this, we train our \alg{} agent with a 2-objective augmentation of the QA dataset, and subsequently test it against other models, using held-out multi-objective test datasets similarly augmented from the original test datasets. 
As elaborated in~\cref{sec:task-design}, these multi-objective tasks require a significantly larger number of turns of environment interactions to complete, hence serving as better benchmarks for memory management.
As shown in~\cref{tab:multi_dataset_metrics}, when evaluated on 2-objective datasets, \alg{} achieves better performance (in terms of EM and F1 scores) than other 7B counterparts, while incurring significantly lower peak token usage and achieving faster inference time.

\begin{figure}[t]
    \centering
    \includegraphics[width=0.9\linewidth]{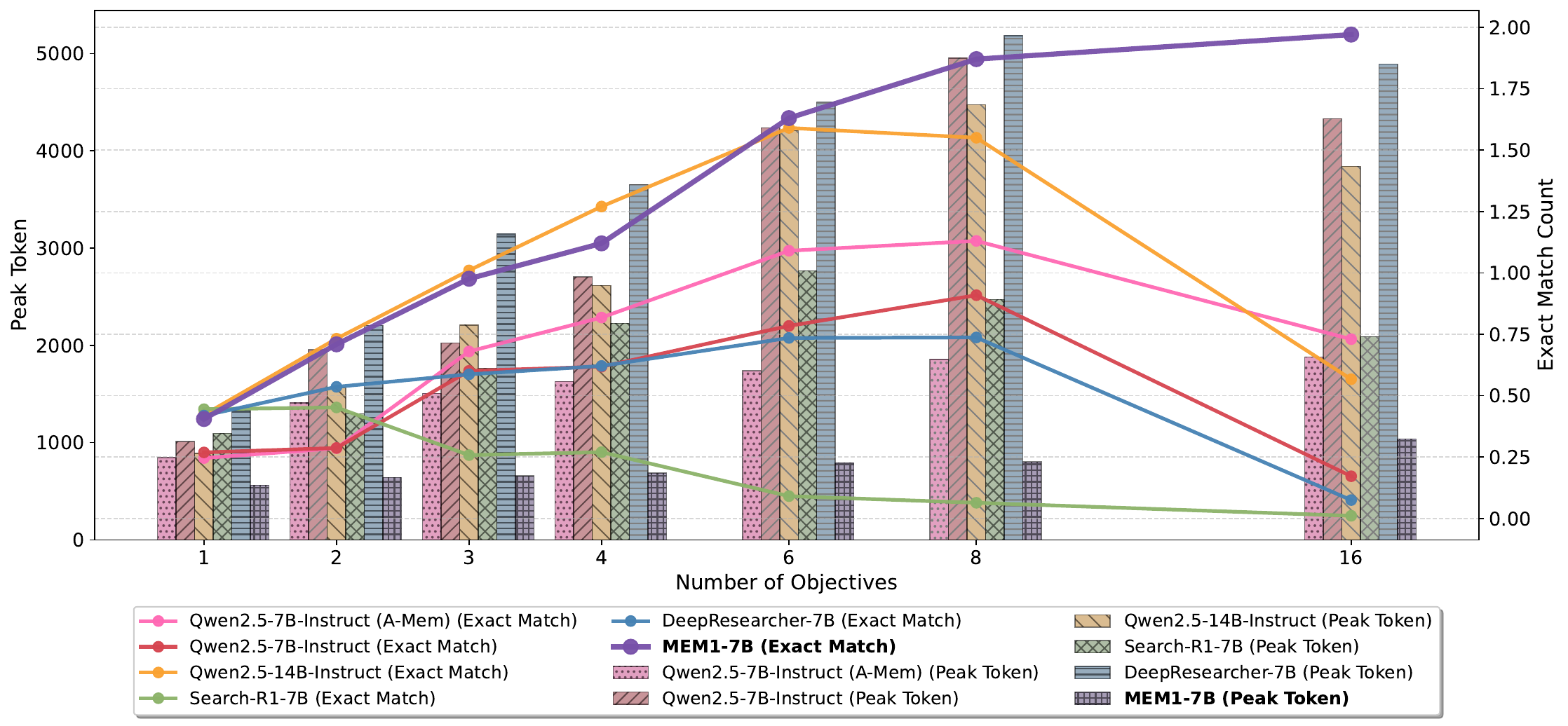}
    \caption{Performance and efficiency scaling of \alg{} (trained on $2$-objective QA) with the number of objectives in multi-objective tasks. \alg{} outperforms the other models and baselines while having an almost constant scaling in memory usage. Note that at $16$-objective, the context of baseline models does not increase anymore since their model performance has degraded (some collapsed).}
    \label{fig:scaling}
\end{figure}

The advantage of \alg{} becomes even more evident in tasks requiring longer-horizon interactive processes.
To highlight such scalability of \alg{}, we further compare the models on $3,4,6,8$, and $16$-objective tasks in~\cref{fig:scaling} and~\cref{tab:multi_dataset_metrics}.
\cref{fig:scaling} illustrates the scaling trends of task performance (measured by EM count) and memory efficiency (measured by Peak Token Usage) for \alg{} relative to other models and memory management baselines. 
As the number of objectives increases, the Peak Token Usage of all other methods and models scales nearly linearly.
In contrast, \alg{} maintains an almost constant peak token count with only a slight increase, as also shown in~\cref{tab:multi_dataset_metrics}.

Notably, while \alg{} initially underperforms Qwen2.5-14B-Instruct, its performance gradually catches up as the number of objectives increases, eventually surpassing the 14B model, which has double the parameter count.
\alg{} also demonstrates remarkable efficiency. 
In the $16$-objective task, it requires only $27.1\%$ of the peak tokens and $29.3\%$ of the total inference time compared to Qwen2.5-14B-Instruct.
This efficiency translates to significantly reduced GPU memory requirements and overall computing resource demands.

\begin{table}[t]
  \centering
  \caption{Comparison of models on multi-objective multi-hop QA tasks. Arrows indicate the desired directions. \textcolor{red}{Numbers in red} indicate collapsed model behavior (extremely low performance). (truncate) means using \alg{}'s prompt and rollout pipeline. (A-MEM) means using \alg{}'s prompt and rollout pipeline with A-Mem's external memory module~\citep{xu2025mem}. \alg{}-QA means \alg{} trained on $2$-objective QA task.}
  \resizebox{\textwidth}{!}{%
    \begin{tabular}{@{}l
        *{4}{c}  
        *{4}{c}  
        *{4}{c}  
        @{}}
      \toprule
      \multirow{2}{*}{\textbf{Model}} 
        & \multicolumn{4}{c}{\textbf{$2$-Objective}} 
        & \multicolumn{4}{c}{\textbf{$8$-Objective}} 
        & \multicolumn{4}{c}{\textbf{$16$-Objective}} \\
      \cmidrule(lr){2-5} \cmidrule(lr){6-9} \cmidrule(lr){10-13}
        & EM ↑ & F1 ↑ & Peak ($\times 10^2$) ↓ & Time (s) ↓ 
        & EM ↑ & F1 ↑ & Peak ($\times 10^2$) ↓ & Time (s) ↓ 
        & EM ↑ & F1 ↑ & Peak ($\times 10^2$) ↓ & Time (s) ↓ \\
      \midrule
      Qwen2.5-14B-Inst
        & \textbf{0.732} & \textbf{0.902} & 15.6$\pm $0.19 & 5.49 $\pm$ 0.16 
        & 1.55 & 1.87 & 44.7 $\pm$ 0.37 & 16.2 $\pm$ 0.27
        & 0.567 & 0.703 & 38.4$\pm$0.71 & 29.7$\pm$0.75 \\
      Qwen2.5-7B-Inst
        & 0.268 & 0.366 & 19.6$\pm$0.33 & 4.60$\pm$0.08 
        & 0.87 & 1.10 & 49.5$\pm$0.40 & 13.9$\pm$0.18 
        & 0.165 & 0.213 & 43.3$\pm$0.62 & 15.5$\pm$0.23 \\
      Qwen2.5-7B-Inst (A-MEM)
        & 0.286 & 0.371 & 14.1$\pm$0.10 & 24.6$\pm$0.51 
        & 1.13 & 1.43 & 18.6$\pm$0.10 & 53.7$\pm$1.26 
        & 0.730 & 0.961 & 18.8$\pm$0.14 & 91.2$\pm$2.44 \\
      Qwen2.5-7B-Inst (truncate)
        & 0.262 & 0.336 & 8.28$\pm $0.06 & 5.89$\pm$0.16 
        & 0.97 & 1.23 & 11.8$\pm$0.10& 11.9$\pm$0.20 
        & 0.396 & 0.497 & 13.3$\pm$0.16 & 22.1$\pm$0.60 \\
      Search-R1 
        & 0.452 & 0.531 & 13.0$\pm$0.08& 4.09 $\pm$ 0.23 
        & \textcolor{red}{0.064} & \textcolor{red}{0.08} & \textcolor{red}{24.7 $\pm$ 0.19} & \textcolor{red}{\textbf{4.25}$\pm$\textbf{0.16}}
        & \textcolor{red}{0.009} & \textcolor{red}{0.011} & \textcolor{red}{20.9$\pm$0.03} & \textcolor{red}{\textbf{4.75}$\pm$\textbf{0.18}} \\
      DeepResearcher 
        & 0.536 & 0.650 & 22.0$\pm$0.43 & \textbf{4.01}$\pm$\textbf{0.07}
        & 0.73 & 0.90 & 51.8$\pm$0.35 & 11.3$\pm$0.14 
        & \textcolor{red}{0.071} & \textcolor{red}{0.106} & \textcolor{red}{48.9$\pm$0.66} & \textcolor{red}{15.8$\pm$0.19} \\
      \textbf{\alg{}-QA} 
        & 0.709 & 0.838 & \textbf{6.40}$\pm$\textbf{0.02} & 6.49 $\pm$ 0.07
        & \textbf{1.87} & \textbf{2.31} & \textbf{8.01}$\pm$\textbf{0.06} & \textbf{8.68}$\pm$\textbf{0.12}
        & \textbf{1.97} & \textbf{2.39} & \textbf{10.4$\pm$0.09} & \textbf{8.70$\pm$0.12} \\
      \bottomrule
    \end{tabular}%
  }
\label{tab:multi_dataset_metrics}
\end{table}


\subsection{\alg{} on Single-Objective Multi-Hop Tasks}

While \alg{} is designed to train agents for very long-horizon tasks, our training method also delivers improved capability with existing multi-hop tasks while achieving much greater efficiency at the same time, all without being explicitly trained on the single-objective versions of these tasks. Note that single-objective tasks also require multiple turns of interaction to produce the desired output.

\paragraph{Long-horizon web navigation in WebShop.}

Beyond QA tasks, we further evaluate the effectiveness of \alg{} in managing long-horizon interactions in the form of web navigation.
We show the experimental results in~\cref{tab:performance-webagent}.
Trained in the WebShop environment (see \cref{app:webshop}), \alg{} outperforms other agent training baselines, including Agent-Flan, Agent-R, and AgentLM when utilizing models of similar size.
Furthermore, \alg{} achieves remarkable efficiency improvements compared to the best baseline method, AgentLM, featuring a $2.8\times$ improvement in Peak Token Usage, a $1.9\times$ improvement in Dependency, and a $1.5\times$ improvement in Inference Time.
\alg{} even surpasses AgentLM-13B, a model with twice the parameter count of our trained model.
Additionally, our results indicate that using \alg{} is significantly better than OpenAI's GPT-4o on the WebShop tasks, even when the truncation prompt templates or A-MEM techniques are applied to GPT-4o.

\begin{table}[t]
  \centering
  \caption{The experimental results for WebShop. For a fair comparison, we do not report GPT's inference time. For Agent-R, scores are taken from the original paper, as the model is closed source. \alg{}-WebShop means \alg{} trained on WebShop environment.}    
  \resizebox{\textwidth}{!}{%
    \begin{tabular}{@{} l c c c c @{}}
        \toprule
        Model & Avg Final Reward \up &  Peak Token ($\times 10^3$) \down & Dependency ($\times 10^6$) \down & Inference Time Per Traj (s) \down \\
        \midrule
        GPT-4o & 25.48 & $5.30 \pm 1.23$ & $3.99 \pm 1.16$ & \textit{N/A} \\
        GPT-4o (truncate) & 13.82 & $0.99 \pm 0.99$ & $0.81 \pm 0.23$ & \textit{N/A} \\
        GPT-4o (A-MEM) & 24.50 & $1.84 \pm 0.06$ & $0.31 \pm 0.11$ & \textit{N/A} \\
        \midrule
        Qwen2.5-7B-Instruct & 18.42 & $5.64 \pm 1.34$ & $3.38 \pm 0.89$ & $12.31 \pm 1.82$ \\
        Qwen2.5-14B-Instruct & 12.34 & $5.44 \pm 0.92$ & $3.30 \pm 0.61$ & $18.17 \pm 2.32$ \\
        Agent-FLAN-7B & 40.35 & $3.37 \pm 1.12$ & $2.18 \pm 1.62$ & $9.95 \pm 6.19$ \\
        Agent-R-8B & 63.91 & \textit{N/A} & \textit{N/A} & \textit{N/A} \\
        AgentLM-7B & 63.60 & $2.24 \pm 0.40$ & $0.28 \pm 0.07$ & $3.91 \pm 1.07$ \\
        AgentLM-13B & 70.80 & $2.36 \pm 0.46$ & $0.30 \pm 0.08$ & $5.23 \pm 1.59$ \\
        \midrule
        \best{\alg{}-WebShop} & \best{70.87} & \best{\pmstd{0.81}{0.10}} & \best{\pmstd{0.15}{0.16}} & \best{\pmstd{2.61}{0.48}} \\
        \bottomrule
    \end{tabular}%
  }
  \label{tab:performance-webagent}
\end{table}

\paragraph{Single-objective QA in Wikipedia.}
\cref{tab:performance-rag} presents the accuracy and efficiency metrics for evaluations on single-objective QA tasks on Wikipedia~\citep{jin2025search}, where the agent can make retrieval requests from the Wikipedia datastore via RAG. The \alg{} used in this evaluation is the same as the one detailed in ~\cref{sec:exp-impl-details:multi-obj}, which is trained solely on a 2-objective task. Overall, \alg{} demonstrates superior efficiency across all three evaluated efficiency metrics, while simultaneously achieving the highest EM score and an F1 score comparable to that of Qwen2.5-14B-Instruct. This improvement in efficiency is attributed to the \alg{} agent's ability to consolidate memory from previous interactions into a compact internal state, which reduces the number of tokens used in the context. We also observe that SFT significantly underperforms RL, highlighting the necessity for RL-based training. 

\begin{table}[t]
  \centering
  \caption{Performance comparison across environments for single-objective tasks. Arrows indicate the desired direction. (SFT) means training with SFT and applying \alg{}'s prompt and rollout. Note that DeepResearcher is specifically trained on the single-objective Online Web-QA task with F1 score as the optimization objective, and Search-R1 is specifically trained on the single-objective Wiki-RAG task with EM as the objective. }
  \resizebox{\textwidth}{!}{%
    \begin{tabular}{@{}ll
        cc
        ccc@{}}
      \toprule
      \textbf{Environment} & \textbf{System} 
        & \makecell{EM \textuparrow} 
        & \makecell{F1 \textuparrow} 
        & \makecell{Peak Token ($\times 10^2$) \textdownarrow} 
        & \makecell{Dependency ($\times 10^5$) \textdownarrow} 
        & \makecell{Inference Time \textdownarrow} \\
      \midrule
      \multirow{6}{*}{Wiki RAG}
      & \makecell[l]{Qwen2.5-7B-Inst (truncate)}    & 0.287 & 0.382 & $6.28 \pm 0.05$   & $1.65 \pm 0.04$ & $2.26 \pm 0.04$ \\
      & \makecell[l]{Qwen2.5-7B-Inst (A-MEM)}    & 0.246 & 0.373 & $8.47 \pm 0.12$   & $0.92 \pm 0.03$ & $11.2 \pm 0.40$ \\
        & Qwen2.5-7B-Inst        & 0.269 & 0.390 & $9.32 \pm 0.19$   & $1.17 \pm 0.04$ & $2.31 \pm 0.04$ \\
        & Qwen2.5-14B-Inst       & 0.422 & \textbf{0.534} & $8.89 \pm 0.21$   & $2.22 \pm 0.10$ & $6.73 \pm 0.24$ \\
        & Search-R1             & \textbf{0.445} & 0.516 & $11.0 \pm 0.25$   & $1.50 \pm 0.05$ & $\mathbf{2.23 \pm 0.14}$ \\
        & DeepResearcher        & 0.419 & 0.503 & $13.3 \pm 0.34$   & $7.04 \pm 0.33$ & $3.86 \pm 0.09$ \\
        & \textbf{\alg{}-QA (SFT)}   & 0.302 & 0.358 & $6.54 \pm 0.05$   & $3.30 \pm 0.13$ & $4.84 \pm 0.21$ \\
        & \textbf{\alg{}-QA}    & 0.405 & 0.471 & $\mathbf{5.63 \pm 0.03}$ & $\mathbf{0.76 \pm 0.02}$ & $3.79 \pm 0.07$ \\
      \midrule
      \multirow{3}{*}{Online Web-QA}
        & Qwen2.5-7B-Inst        & 0.334 & 0.451 & $8.37 \pm 0.18$ & $1.39 \pm 0.06$ & $2.20 \pm 0.04$ \\
        & DeepResearcher        & 0.372 & \textbf{0.492} & $10.27 \pm 0.19$ & $2.86 \pm 0.14$ & $2.87 \pm 0.06$ \\
        & \textbf{\alg{}-QA}    & \textbf{0.397} & 0.485 & \best{\pmstd{5.79}{0.06}} & \best{\pmstd{0.44}{0.02}} & \best{\pmstd{1.84}{0.03}} \\
      \bottomrule
    \end{tabular}%
  }
  \label{tab:performance-rag}
\end{table}

\paragraph{Zero-shot transfer to Online Web-QA.}
To validate the transferability and generalizability of the trained \alg{} agent, we perform a zero-shot transfer to an online web-QA environment, which is unseen by the agent. 
In this environment, agents conduct web searches through an API service that returns results including titles, snippets, and URLs.
As shown in~\cref{tab:performance-rag}, \alg{} consistently exhibited improved efficiency alongside comparable effectiveness in this unseen setting via zero-shot transfer.

\subsection{Analysis on Emergent Agent Behaviors}
\label{sec:analysis}
{\setlength{\fboxsep}{0pt}%
\begin{figure}[t]
    \centering
    \includegraphics[width=\linewidth]{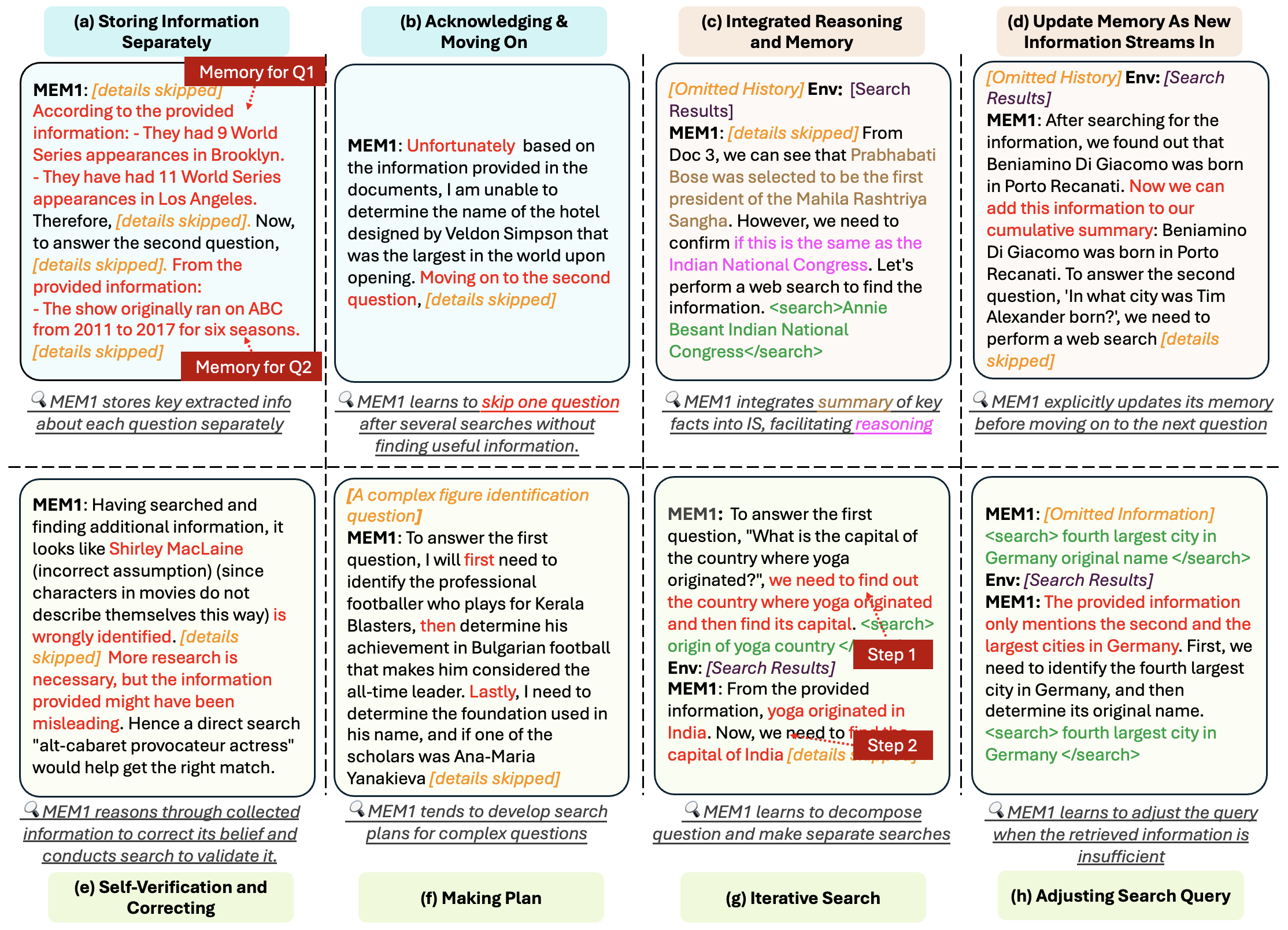}
    \caption{Snippets of internal states and actions showing \alg{}'s Emergent Behaviors in 2-objective QA tasks. \colorbox{multibehaviorcolor}{Light Blue} denotes behaviors related to multi-objective tasks. \colorbox{memorybehaviorcolor}{Beige} denotes behaviors related to memory in internal state. \colorbox{searchbehaviorcolor}{Pastel Green} denotes behaviors related to general search strategies.}
    \label{fig:model_behavior}
\end{figure}

Through analyzing \alg{}’s multi-turn interaction traces trained on 2-objective QA, we observe a range of emergent behaviors that are critical for handling long-horizon, multi-objective tasks, demonstrating capabilities well beyond simple retrieval. First, \alg{} learns to \textbf{manage multiple questions concurrently} by maintaining a structured internal state. As shown in~\cref{fig:model_behavior}(a), when faced with two multi-hop questions, the agent stores and updates memory for each question separately, guiding subsequent searches based on the identified information gaps. In (b), \alg{} exhibits the ability to shift focus when progress on one question stalls, recognizing difficulty and prioritizing the more tractable objective. Meanwhile, \alg{} learns to \textbf{interleave reasoning and memory} in its internal state $\colortag{thinkcolor}{<IS>}$, weaving important information into its decision-making process to support both information retention and action selection. In \cref{fig:model_behavior} (c), \alg{} explicitly extracts important information from previous search results and leverages it to formulate the next query that best addresses the current information gap. In addition, (d) shows that when new, relevant information is retrieved, \alg{} explicitly reasons about its significance and selectively updates its memory. We believe that learning these interleaved behaviors is key to achieving efficiency gains in memory without degrading performance. 
Beyond behaviors unique to our multi-objective setup and memory architecture, \alg{} also exhibits several \textbf{general-purpose search strategies}. In (e), the agent performs self-verification, correcting an earlier misconception and issuing a new query for confirmation. In (f), complex queries are decomposed into manageable subgoals before initiating the search. In (g), for questions requiring multi-turn information gathering, \alg{} extracts key information from search results and uses it to inform the next search. In (h), when overly specific queries fail, \alg{} re-scopes its query to improve retrieval. Notably, many of these behaviors, including verification, making a plan, and iterative search, are also reported in recent studies on deep research agents \cite{jin2025search,zheng2025deepresearcher}.

\section{Conclusion, Limitations, and Future Work}
\label{sec:conclusion}

We introduced \alg{}, a reinforcement learning framework that enables language agents to perform long-horizon reasoning with consolidated memory. By integrating inference-time reasoning and memory consolidation into a unified internal state, \alg{} addresses the scalability challenges of prompt growth and achieves competitive performance across QA and web navigation benchmarks, with substantially reduced memory usage and inference latency. Despite these advantages, \alg{} assumes access to environments with well-defined and verifiable rewards. While this assumption holds in domains such as QA, math, and web navigation, many open-ended tasks present ambiguous or noisy reward structures. Fully realizing the potential of \alg{} therefore requires advances in modeling such tasks and designing suitable reward mechanisms—challenges that lie beyond the scope of this work. A promising future direction is to explore methods for training \alg{} agents in these open-ended settings where reward signals are sparse, delayed, or implicit.

\newpage
\bibliographystyle{natbib}
\bibliography{reference}

\newpage
\appendix
\section{Details of \alg{}}

\subsection{Computing Resources and Training Details}
\label{app:compute-rsc}

All trainings of \alg{} are conducted on $4$ H100 or H200 GPUs. We use the veRL framework~\citep{sheng2024hybridflow} for RL and Swift~\citep{swift2024} for SFT. For RL, both the data batch size and mini batch size are set to $64$. Learning rate is set to $10^{-6}$ for the actor model and $10^{-5}$ for the critic model with a linear warmup of $50$ steps. Temperature is set to $1$ during training and $0.01$ during inference.

All evaluations are conducted on a single H200 GPU, which serves the respective models as an API service using the vLLM framework~\citep{kwon2023efficient} with automatic prefix caching enabled.

\subsection{RAG Configuration}

For RAG on local Wiki corpus, we use Faiss-GPU~\citep{douze2024faiss} serving an E5 Base model~\citep{wang2022text}. The Wiki corpus is taken from a Wikipedia 2018 dump~\citep{karpukhin-etal-2020-dense}. The number of passages for each retrieval is set to $3$ for a fair comparison with other methods.

For online web search queries, we use Serper API~\citep{serper2025}, which offers Google search results including titles, snippets, and URLs. For each search, we return the top $10$ results to the agent as external information. We do not ask the agent to retrieve the content of specific webpages.

\subsection{Prompts}\label{app:prompts}

\begin{mycolorbox}[label=multi-prompt]{Prompt}{Multi-Objective Task (QA)}{
\texttt{You will answer multiple complex questions using iterative reasoning, summarization, and web search. \\\\
At each step, you will see the questions, a cumulative summary of relevant information, the current search query, and search results (except in the first step, where only the questions are provided). Your task is to: \\\\
1. Perform reasoning and update a cumulative, concise summary within <think> ... </think>. This acts as persistent memory and must include all essential information from previous <think> and <information> tags.\\\\
2. Then choose one of the following actions:\\
- If any question remains unanswered, issue a single query for one question inside <search> ... </search>. The query should consist of keywords or a short phrase. Only search one question at a time.\\
- If all questions are answered, provide the final answers—separated by semicolons—within <answer> answer1; answer2; ... </answer>. The answers must be concise, contain only essential words, and avoid any explanations.\\\\
Important:\\
- Always follow this structure after <information> or the initial questions: <think> ... </think><search> ... </search> or <think> ... </think><answer> ... </answer>.\\
- Do not search multiple queries or questions simultaneously.\\\\
Answer the following questions:[QUESTIONS]}}
\end{mycolorbox}

\begin{mycolorbox}[label=single-prompt]{Prompt}{Single-Objective Task (QA)}{
\texttt{You will answer a complex question through iterative reasoning, summarization, and web searches.\\\\
At each step, you can see the question, previous summary in <think> ... </think>, search query in <search> ... </search>, and the returned information in <information> ... </information> (except the first step where you will be given only the question). Then, you should:\\\\
1. Conduct reasoning, and then update a concise, cumulative summary with essential information inside <think> </think>. This is your persistent memory and should include all important information from previous <think> </think> and <information> </information> (i.e. information and answers already found for questions).\\\\
2. Then choose one:\\
- Issue a query (i.e., key words / phrases for search) inside <search> </search> (you may search repeatedly until the answer is clear). This query will be used to conduct search and return the results in <information> results </information>\\
- Provide the final concise answer (no explanations) if no additional information is needed inside <answer> </answer>. The answer should be concise and only contain the words necessary to answer the question.\\\\
After <information> </information> (or question at the beginning), you should always follow the order: <think> ... </think><search> ... </search> or <think> ... </think><answer> ... </answer>.\\\\
Question: [QUESTION]}}
\end{mycolorbox}

\begin{mycolorbox}[label=single-prompt-webshop]{Prompt}{Single-Objective Task (WebShop)}{
\texttt{You are browsing an online shop. Your goal is to find a product that matches the given description. You will interact with the site step-by-step. Each step gives you a <state>...</state> representing the current webpage. You must decide what action to take next until you identify the correct product.
\\\\
Available actions (shown in the <state> tag) depend on the page: \\
- On the search page: search[<keywords>] \\
- On search result pages: click[<item url>] to view a product, or click[next >] to go to the next results page \\
- On product pages: click[description], click[features], click[color], click[size], click[buy now] \\
- To return to search: click[back to search]
\\\\
Example goal: "Find a gingko light and 20x20 pillow cover that is hand painted." 
Example first action: <answer>search[gingko light 20x20 pillow cover hand painted]</answer>
Only respond with valid actions formatted as: search[...], click[...], etc.
\\\\
After you navigate and find the product that best fits the user goal, you should click[buy now] to buy the product at the product page when the buy now button is available.\\\\
Product Description: [PRODUCT DESCRIPTION]}}
\end{mycolorbox}

\subsection{Implementation Details of Metrics and Baselines}

\subsubsection{Metrics}\label{app:metrics}

\paragraph{Exact match.}
In QA tasks, we use exact match (EM) as both the verifiable reward for the RL pipeline and the evaluation metric for the final output.
The final response is extracted from between
$\colortag{answercolor}{\texttt{<answer>}}$ and $\colortag{answercolor}{\texttt{</answer>}}$. 
In multi-objective settings, the response should contain answers to each question separated by semicolons. 
If the XML tags are mismatched, or if the number of provided answers does not correspond to the number of questions, a score of $0$ is assigned. 
Otherwise, $1$ point is credited for each correct answer.
During RL training, we do not provide any other intermediate rewards or format penalties, as we find that such manual interventions can interfere with the agent's learning process (see more in \cref{sec:analysis}).

\paragraph{F1 score.} The F1 score computes the harmonic mean between the precision $p$ and recall $r$. In the case of string matching, we split both the predicted answer and the ground truth. For example, if the ground truth is ``United States of America'', it is split into a list with lower-case words: ``united'', ``states'', ``of'', ``america''. The same works for the predicted answer. Then, denote the number of common words as $c$.  Further denote the number of words in the predicted answer as $l$ and the number of words in the ground truth as $g$. Then, precision is calculated as $p \coloneqq c / l$ and recall is calculated as $r \coloneqq c / g$. The F1 score is finally computed as
\begin{equation*}
    \texttt{F1} \coloneqq 2\times \frac{p \times r}{p + r} \ .
\end{equation*}
If multiple ground truths are present, the maximum of all F1 scores is chosen. For multi-objective tasks, the final F1 is the sum of the F1 scores for each sub-question.

\paragraph{Peak token usage.} Peak token usage is calculated as the maximum number of tokens (using GPT-4o-mini tokenizer) in any single sequence throughout the agent's entire trajectory.
For fair comparison in our experiments, the system prompt is excluded when computing this sequence length.
The peak token usage serves as a proxy for the inference-time memory requirement.

\paragraph{Dependency length.} Following~\citep{zhang2025lightthinker}, the dependency metric is defined as the total number of historical tokens on which each generated token effectively depends.
Let $T$ denote the total number of interaction steps.
For each step $i \in [T]$, let $n_p^{(i)}$ be the number of prefix tokens and $n_o^{(i)}$ be the number of output tokens generated.
The dependency metric is then calculated as
\begin{equation*}
    \texttt{Dependency} \coloneqq \sum_{i \in [T]} \frac{(2n_o^{(i)}+n_p^{(i)}) \times n_o^{(i)}}{2} \ .
\end{equation*}
At a high level, this metric quantifies the cumulative computational cost associated with the generation of an output trajectory.
It is important to note that in \alg{}, prefix tokens from previous steps are consolidated into a new internal state, rather than being continuously accumulated.
In our experiments, we ignore the tokens in the system prompt when calculating the dependency metric. 


\paragraph{Inference time.} Inference time for each trajectory is recorded as the total elapsed time required to generate the complete output trajectory.
For all experiments, these measurements are conducted on a single H200 GPU, operating with 10 concurrent threads.
The vLLM inference framework is utilized, with its automatic prefix caching feature enabled.

\subsubsection{Baselines}\label{app:baselines}

\paragraph{Search-R1.} As detailed in~\citep{jin2025search}, the model is trained on the $1$-objective task with the same dataset as \alg{}. Search-R1 also uses exact match as its reward function. In comparison, \alg{} is trained exclusively on $2$-objective tasks.

\paragraph{Deep Researcher.} As detailed in~\citep{zheng2025deepresearcher}, the model is trained on $1$-objective task with a curated set from various QA datasets including HotPotQA and Natural Questions. Deep Researcher adopts the F1 score as the reward function.

\subsection{Algorithm}\label{app:algo}

We provide an outline of the rollout of \alg{}, which actively manages its context in \cref{alg:mem1-rollout}. Parts of the pseudo-code follow~\citep{jin2025search}.

\begin{algorithm}[H]
\caption{\alg{} Rollout}
\begin{algorithmic}[1]
\label{alg:mem1-rollout}
\REQUIRE Task prompt $x$, policy model $\pi_\theta$, world model $\mathcal{W}$, maximum turn $T$
\ENSURE Final response $y$
\STATE Initialize rollout sequence $y \leftarrow \varnothing$
\STATE Initialize turn count $t \leftarrow 0$
\WHILE{$t < T$}
    \STATE Initialize current policy rollout sequence $y_t \leftarrow \varnothing$
    \WHILE{True}
        \STATE Generate response token $y_r \sim \pi_\theta(\cdot \mid x, y + y_t)$
        \STATE Append $y_r$ to rollout sequence $y_t \leftarrow y_t + y_r$
        \IF{($t = T - 1$) and $y_r \in [\colortag{answercolor}{\texttt{</answer>}}, \texttt{<eos>}]$}
            \STATE \textbf{break} \hfill {\color{blue} // prevent the agent from searching further}
        \ELSIF{$y_r \in [\colortag{querycolor}{\texttt{</query>}}, \colortag{answercolor}{\texttt{</answer>}}, \texttt{<eos>}]$}
            \STATE \textbf{break}
        \ENDIF
    \ENDWHILE
    \STATE $y \leftarrow y_t$ \hfill {\color{blue} // all previous context removed.}
    \IF{\colortag{querycolor}{\texttt{<query>}} \colortag{querycolor}{\texttt{</query>}} detected in $y_t$}
        \STATE Extract search query $q \leftarrow \text{Parse}(y_t, \colortag{querycolor}{\texttt{<query>}},\colortag{querycolor}{\texttt{</query>}})$
        \STATE Retrieve environment feedback $d \leftarrow \mathcal{W}(q)$ from local storage, Search engine, HTML, $\cdots$
        \STATE $\texttt{HINT} \gets \texttt{You have \{$T - t$\} turns left.}$
        \STATE Insert $d$ into rollout $y \leftarrow y + \colortag{infocolor}{\texttt{<info>}}\texttt{HINT} + d \colortag{infocolor}{\texttt{</info>}}$
    \ELSIF{\colortag{answercolor}{\texttt{<answer>}} \colortag{answercolor}{\texttt{</answer>}} detected in $y_t$}
        \RETURN final generated response $y$
    \ELSE
        \STATE Mark the sample as invalid
    \ENDIF
    \STATE Increment turn count $t \leftarrow t + 1$
\ENDWHILE
\RETURN final generated response $y$
\end{algorithmic}
\end{algorithm}

\subsection{\alg{} on Webshop Training Details}
\label{app:webshop}

We use the same rollout pipeline and policy update mechanism for training \alg{} on WebShop. Compared to the QA tasks, we use a tailored prompt that retains the gist of memory consolidation with instructions specific to the WebShop environment, as shown in Prompt.~\ref{single-prompt-webshop}. Another distinction is that the WebShop environment comes with its own reward function corresponding to each state. Therefore, we do not use exact match but the built-in reward function as the reward signal when training in WebShop environment. The training and test splits also follow the original paper~\citep{yao2022webshop}, with the first $1000$ samples as the test set, the $1000$th to $1500$th as the val set, and the remaining as the train set.

\subsection{Additional Discussion on the Attention Matrix Design.}

We wish to note that our modification to the attention matrix \textit{does not} fully recover the attention of the original trajectories because of the change in position ids. Specifically, prior works~\citep{leviathan2023fast,chen2023accelerating,cai2024medusa} that utilized the attention matrix to compress multiple trajectories mainly targeted tree-exploration, i.e., generating multiple sequences with the same prefix. For these works, on top of the attention matrix, they adjusted the position ids as well, so each trajectory follows a consecutive increasing position ids. However, in MEM1, the prefix does not remain the same because of memory consolidation. This results in each \texttt{<IS>} having two possible position ids, one for the previous turn and one for the next turn. To completely recover the original attention, we need to duplicate each \texttt{<IS>} and assign different position ids to the two copies. However, such duplication can significantly slow down training because the training trajectories are now much longer.

As such, for training efficiency, we do not duplicate the \texttt{<IS>} and assign the position ids for the previous trajectory to each \texttt{<IS>}. While this modification slightly deviates from the ``ideal'' implementation, effectively, it can be viewed as simply adding white spaces in the training trajectories and has no significant impact on the experimental results.

\section{Broader Impacts}
\label{app:broader-impacts}

\alg{} opens up the potential to enable more scalable, efficient, and intelligent AI agents capable of sustaining long, goal-directed interactions in dynamic environments. As AI systems are increasingly deployed in complex real-world tasks—such as scientific research, legal analysis, personalized education, and digital customer service—models must go beyond single-turn capabilities and manage evolving contexts over many steps. \alg{}’s memory-consolidation mechanism allows language models to maintain high performance without the growing computational and environmental costs typically associated with long-context processing. By reducing inference-time memory and compute demands, \alg{} paves the way for more sustainable and scalable AI deployment, making advanced reasoning agents accessible to a wider range of users and institutions, including those with limited resources. Moreover, \alg{}'s unified framework of reasoning and context consolidation sets a precedent for future research on intelligence that can learn to adapt, reflect, and summarize information autonomously, inspiring more trustworthy, interpretable, and human-aligned AI systems.

\section{Training Trajectory Analysis of \alg{}}

\begin{figure}[ht!]
    \centering
    \begin{minipage}[b]{0.48\linewidth}
        \centering
        \includegraphics[width=\linewidth]{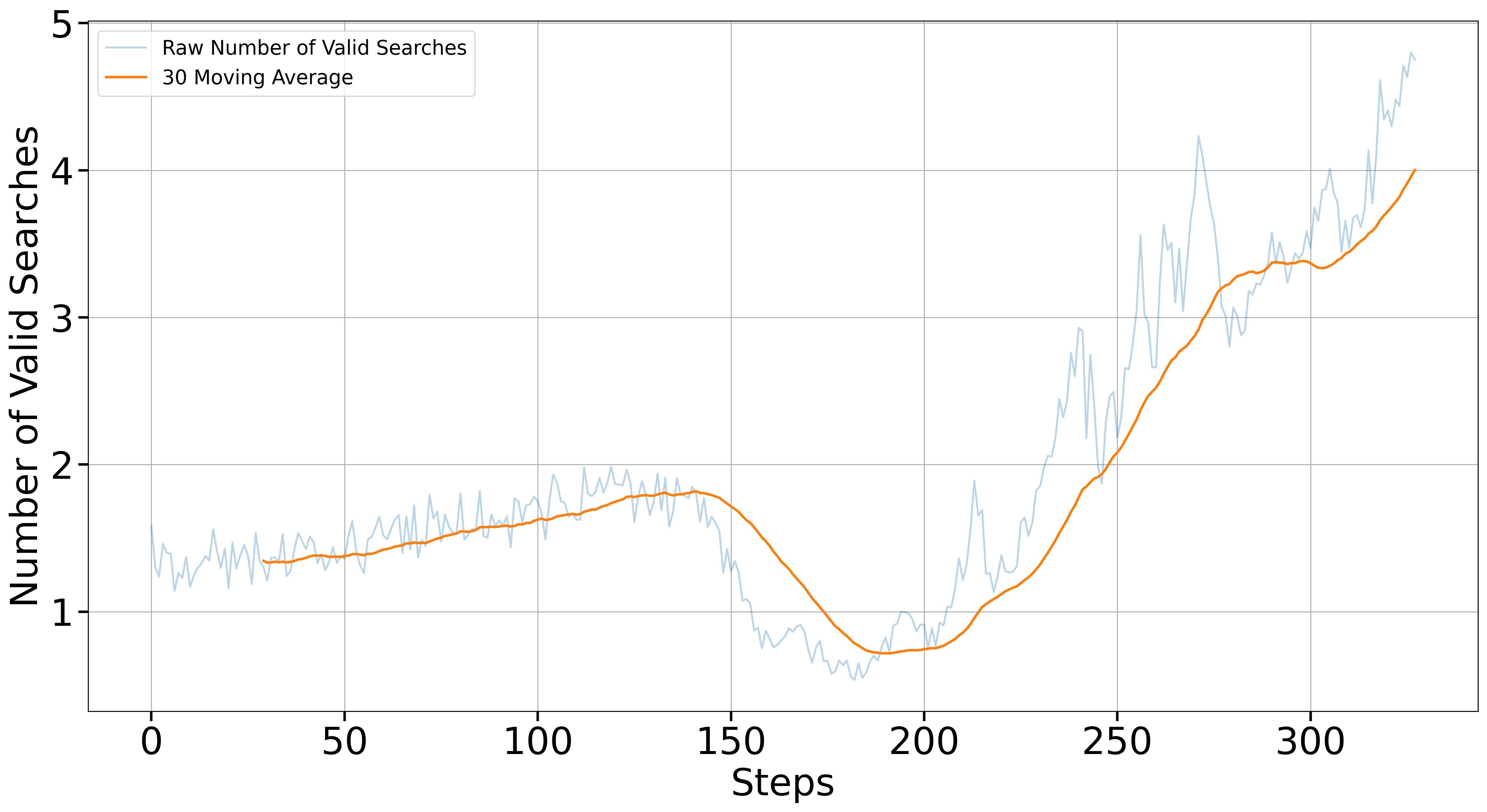}
    \end{minipage}%
    \hfill
    \begin{minipage}[b]{0.48\linewidth}
        \centering
        \includegraphics[width=\linewidth]{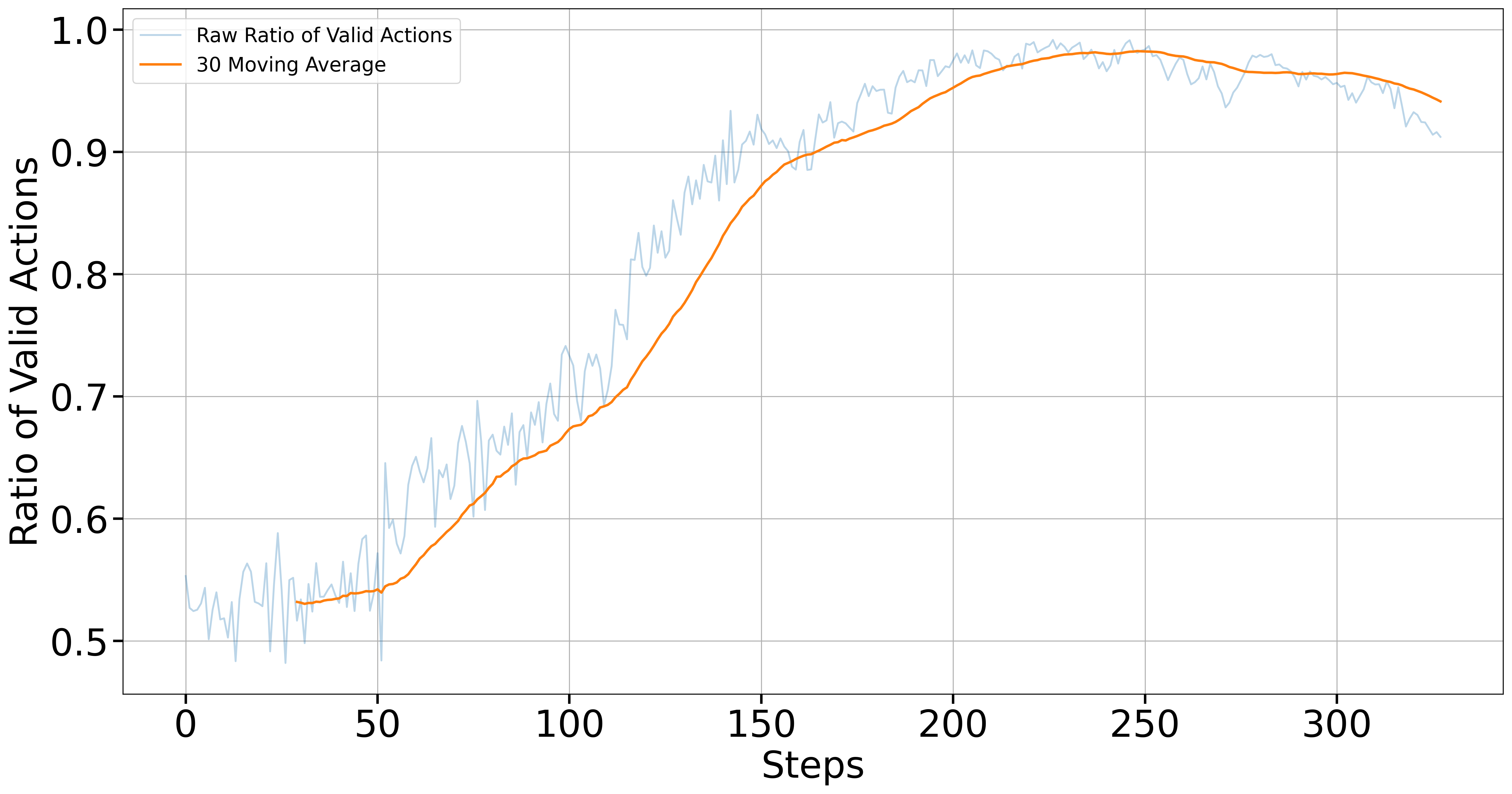}
    \end{minipage}

    \vspace{0.4em}

    \begin{minipage}[b]{0.48\linewidth}
        \centering
        \includegraphics[width=\linewidth]{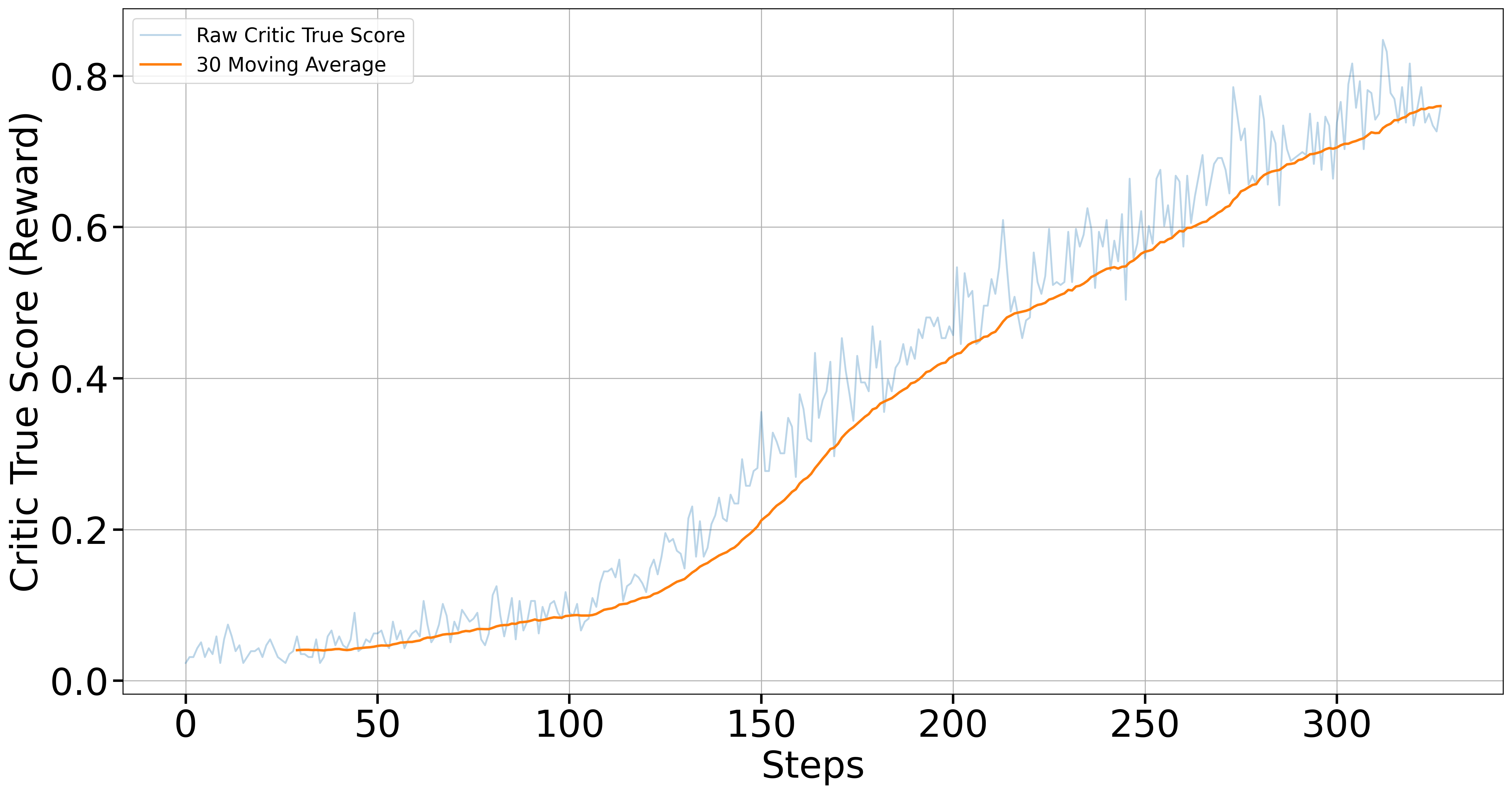}
    \end{minipage}%
    \hfill
    \begin{minipage}[b]{0.48\linewidth}
        \centering
        \includegraphics[width=\linewidth]{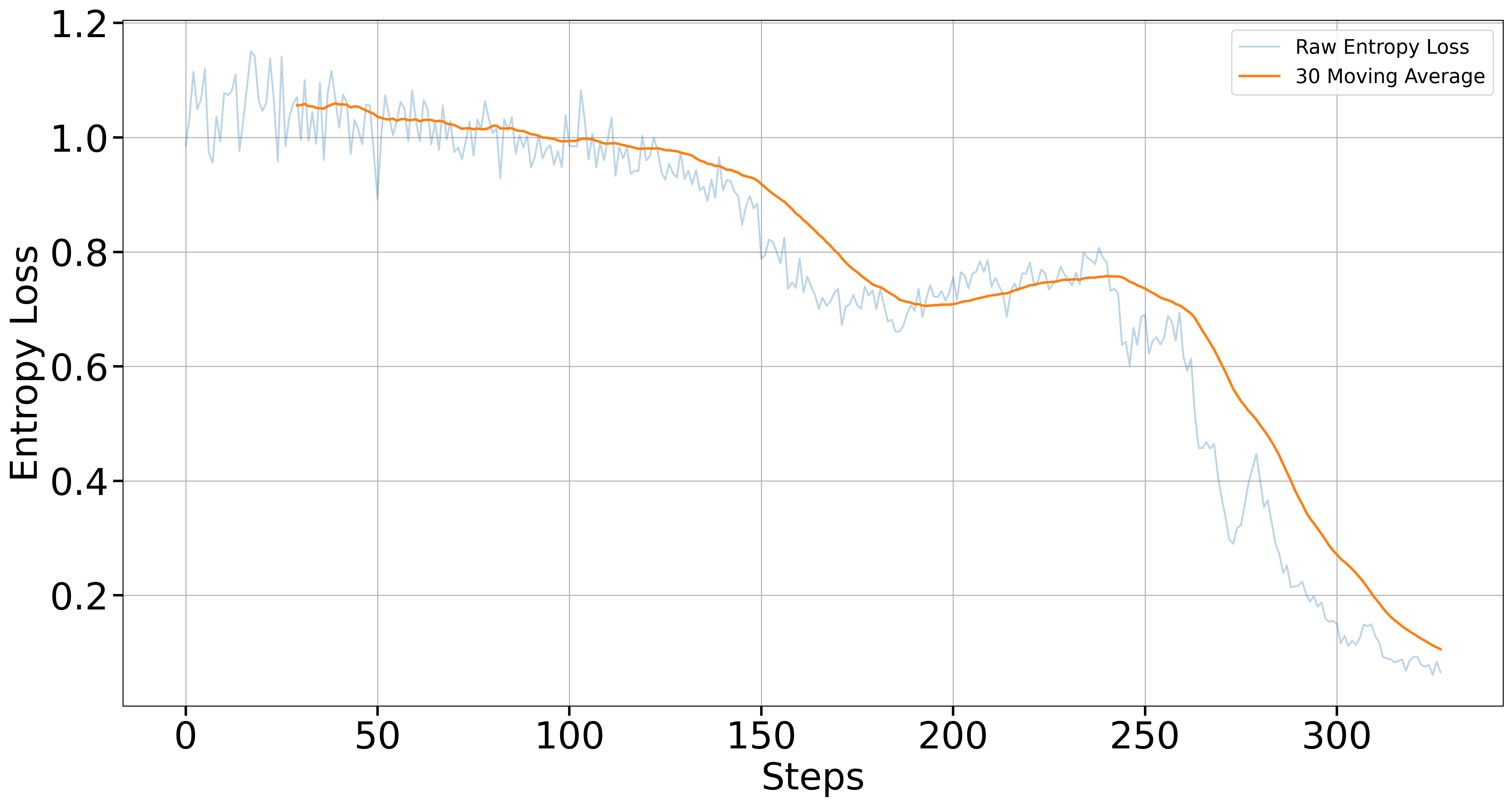}
    \end{minipage}

    \caption{Metrics of training progresses for \alg{} with RL.}
    \label{fig:agent-metrics}
\end{figure}

We present the training dynamics of the 2-objective QA-trained \alg{} in \cref{fig:agent-metrics}, where several distinct phases emerge during the learning process. In the initial exploration phase (first 50 steps), the agent demonstrates little task proficiency. The reward remains consistently low, while the entropy loss is high, suggesting random or undirected behavior. The ratio of valid actions hovers around 0.55, indicating that the agent frequently fails to follow the expected output format. During this period, \alg{} has not yet learned to reliably use the required structure involving $\colortag{querycolor}{<query>}$ and $\colortag{answercolor}{<answer>}$ tags.

Shortly after, we observe the onset of format acquisition. The agent gradually improves its structural consistency, reflected in the rising ratio of valid actions. This improved adherence to format correlates with an increase in reward, suggesting that proper formatting directly contributes to the agent's task success. By around step 150, a notable behavioral shift occurs. The number of valid searches begins to drop sharply, while the reward continues to increase. This implies that the agent has discovered a shortcut: by reducing the number of searches—perhaps to avoid format violations—it can maintain high format fidelity and improve its reward without fully solving the task. This short-horizon optimization suggests the agent is exploiting the reward structure, favoring formatting compliance over content completeness.

Between steps 150 and 200, the agent enters a phase of refined format mastery. The ratio of valid actions steadily climbs, but the number of searches remains low. During this phase, reward growth slows, and entropy begins to flatten. The plateau in entropy indicates that the agent is looking for new policies to boost the reward. At this stage, the agent has reached a local optimum: it's producing valid but under-informed answers.

After step 200, a second behavioral shift occurs. The number of valid searches begins to rise again, suggesting that the agent is learning to extend its interaction horizon to gather more information. The agent learns to balance formatting constraints with information acquisition. As a result, the reward increases more sharply. Finally, after step 250, the agent enters a phase of policy consolidation. The entropy loss drops sharply—signaling a transition from exploration to exploitation—as the agent settles into a more deterministic, high-reward policy. By this stage, the agent effectively combines format compliance, sufficient searching, and high-quality answer generation.

\section{Analysis on Implementation Details}
\subsection{RL Generalizes Better Than SFT}
A natural question arises: can Supervised Fine-Tuning (SFT) with high-quality trajectories match the performance of reinforcement learning (RL)? To investigate this, we compare \alg{}-QA trained via RL against \alg{}-QA (SFT), where both models are trained on the 2-objective QA task. Additionally, the SFT model is further trained on 1-objective and 3-objective QA tasks to enhance its generalization ability. As shown in \cref{tab:rl_sft_generalization}, the SFT model consistently underperforms compared to its RL counterpart across tasks with varying numbers of questions (objectives). Notably, when the number of objectives exceeds six, the performance of the SFT model collapses, whereas the RL-trained model continues to demonstrate strong robustness and scalability.

\begin{figure*}[!ht]
  \centering
  \begin{minipage}{0.48\textwidth}
    \captionof{table}{Comparison of RL and SFT on increasing number of multi-turn questions. Exact match scores ↑ is better. Gap shows absolute difference. Red numbers show collapsed SFT behavior.}
    \centering
    \resizebox{1\linewidth}{!}{
    \begin{tabular}{@{}c|ccc|c@{}}
      \toprule
      \textbf{\#Q} & \textbf{RL ↑} & \textbf{SFT ↑} & \textbf{Gap ↑} & \textbf{RL Gain (\%) ↑} \\
      \midrule
      1  & 0.410 & 0.300 & 0.110 & +36.7\% \\
      2  & 0.709 & 0.433 & 0.276 & +63.7\% \\
      3  & 0.976 & 0.648 & 0.328 & +50.6\% \\
      4  & 1.120 & 0.626 & 0.494 & +78.9\% \\
      6  & 1.630 & \textcolor{red}{0.088} & 1.542 & +1752\% \\
      8  & 1.870 & \textcolor{red}{0.027} & 1.843 & +6826\% \\
      16 & 1.900 & \textcolor{red}{0.000} & 1.900 & --- \\
      \bottomrule
    \end{tabular}}
    \label{tab:rl_sft_generalization}
  \end{minipage}
  \hfill
  \begin{minipage}{0.48\textwidth}
    \centering
    \includegraphics[width=\linewidth]{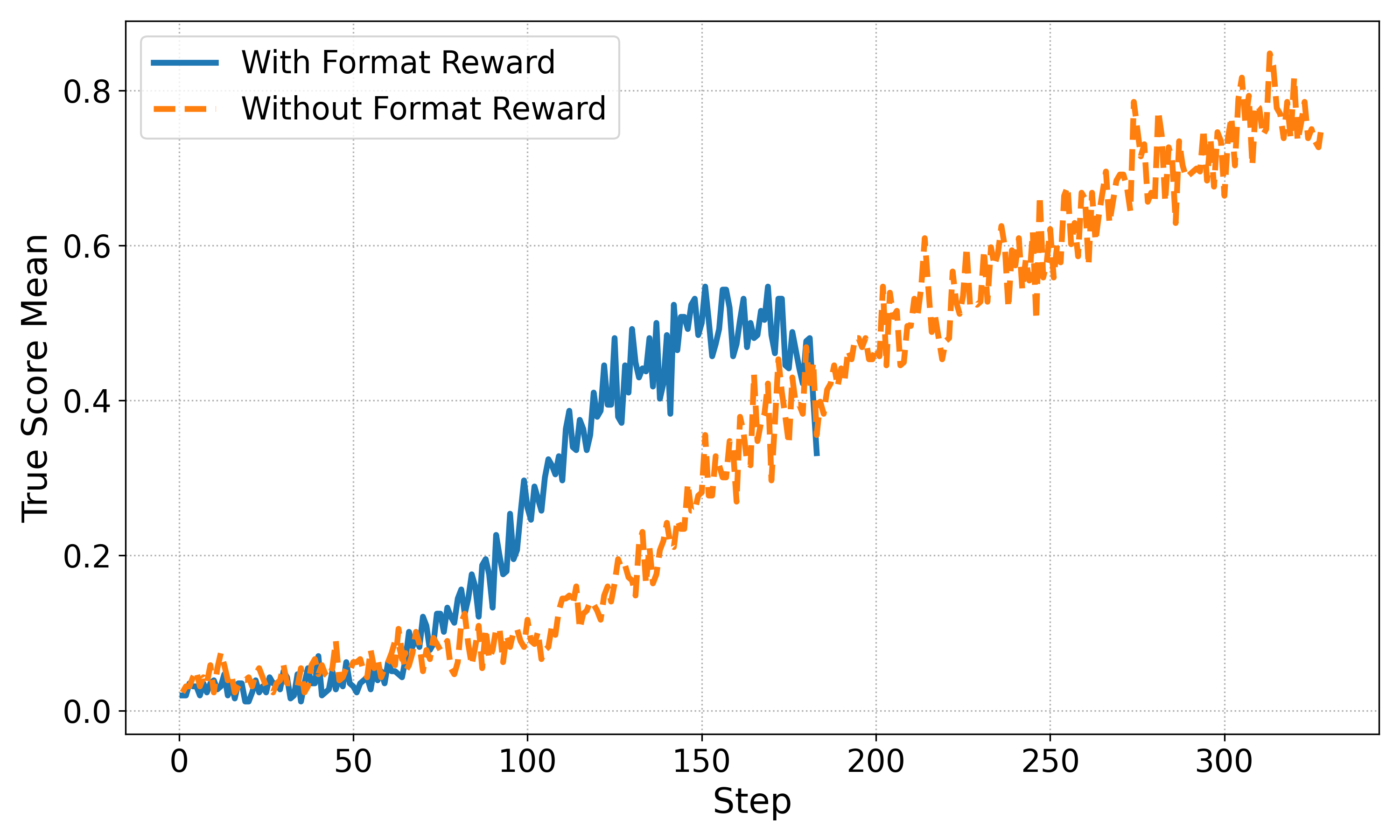}
    \caption{Training curves comparing \alg{} trained with and without format reward.}
    \label{fig:format_reward}
  \end{minipage}
\end{figure*}

\subsection{Format Reward Accelerates Convergence but Degrades Final Performance}
It is common to incorporate format reward when training reasoning models and multi-turn reasoning agents \cite{deepseek2025r1,zheng2025deepresearcher,jin2025search}. In our study, we experimented with a format reward that enforces the agent to produce outputs using specific structural tags: $\colortag{thinkcolor}{\texttt{<IS>}}$, $\colortag{querycolor}{\texttt{<query>}}$, and $\colortag{answercolor}{\texttt{<answer>}}$. If the agent fails to use the expected tags correctly, the turn is terminated and a penalty of -1 is applied.

As shown in \cref{fig:format_reward}, using the format reward leads to faster convergence during training but results in worse final performance. The format-constrained agent achieves an exact match score of 0.466, compared to 0.709 for \alg{} trained with only outcome-based reward on the same testing set for the 2-objective QA task. Additionally, the format-constrained agent generates fewer tokens, with an average peak of 514.9 tokens, whereas the outcome-reward-trained \alg{} reaches an average peak of 640 tokens.

We hypothesize that the format reward accelerates structural learning but constrains exploration of effective reasoning strategies. As a result, the agent learns to produce shorter responses with valid syntax but develops less effective internal state representations, leading to degraded task performance.


\end{document}